\documentclass[journal]{IEEEtran}
\usepackage[numbers]{natbib}
\usepackage{amssymb}
\usepackage{amsmath,graphicx}
\usepackage{subcaption}
\usepackage{booktabs}
\usepackage{adjustbox}
\usepackage[table]{xcolor}
\usepackage[hyphens]{url}
\usepackage{amsfonts}
\usepackage{wrapfig}
\usepackage{graphicx}
\usepackage{mwe}
\usepackage{scrextend}
\usepackage{enumitem}
\usepackage{array}
\usepackage{booktabs}
\usepackage{tabularx}
\usepackage{makecell, soul}

\usepackage[colorlinks=true,urlcolor=blue,linkcolor=red,citecolor=blue]{hyperref}
\usepackage{arydshln}
\usepackage{epsfig}
\usepackage{bbding}
\usepackage{multirow}
\usepackage[table]{xcolor}
\definecolor{grannysmithapple}{rgb}{0.66, 0.89, 0.63}
\definecolor{green(colorwheel)(x11green)}{rgb}{0.0, 1.0, 0.0}
\definecolor{lightgreen}{rgb}{0.56, 0.93, 0.56}
\definecolor{lightcoral}{rgb}{0.94, 0.5, 0.5}
\usepackage{graphicx}

\usepackage{acronym}
\newacro{gan}[GAN]{generative adversarial network}
\newacro{vae}[VAE]{variational autoencoder}
\newacro{cnn}[CNN]{convolutional neural network}
\newacro{dire}[DIRE]{diffusion reconstruction error}
\newacro{mlp}[MLP]{multi-layer perceptron}
\newacro{clip}[CLIP]{contrastive language image pre-training}
\newacro{vlm}[VLM]{vision-language model}
\newacro{vit}[ViT]{vision transformer}
\newacro{nlp}[NLP]{natural language processing}
\newacro{cv}[CV]{computer vision}
\newacro{blip}[BLIP]{bootstrapping language image pre-training}
\newacro{vqa}[VQA]{visual question answering}
\newacro{lora}[LORA]{low-rank adaptation}
\newacro{peft}[PEFT]{parameter-efficient fine-tuning}
\newacro{gpt}[GPT]{generative pre-trained transformer}
\newacro{q-former}[Q-Former]{querying transformer}
\newacro{sedid}[SeDID]{stepwise error for diffusion-generated image detection}
\newacro{sd}[SD]{stable diffusion}
\newacro{lsun}[LSUN]{large-scale scene understanding}
\newacro{fc}[FC]{fully-connected}
\newacro{deit}[DeiT]{data-efficient image transformers}
\newacro{ldm}[LDM]{latent diffusion model}
\newacro{adm}[ADM]{ablated diffusion model}
\newacro{ddpm}[DDPM]{denoising diffusion probabilistic models}
\newacro{iddpm}[IDDPM]{improved denoising diffusion probabilistic models}
\newacro{pndm}[PNDM]{pseudo numerical methods for diffusion models on manifolds}
\newacro{srm}[SRM]{spatial rich model}
\newacro{lasted}[LASTED]{language-guided synthesis detection}
\newacro{rf}[RF]{random forest}
\newacro{dm}[DM]{diffusion model}
\newacro{ddim}[DDIM]{denoising diffusion implicit models}
\newacro{multilid}[multiLID]{multi local intrinsic dimensionality}
\newacro{ifdl}[IFDL]{image forgery detection and localization} 
\newacro{svm}[SVM]{support vector machine} 
\newacro{ai}[AI]{artificial intelligence} 

\newacro{amsff}[AMSFF]{attention-based multi-scale feature fusion} 
\newacro{psm}[PSM]{patch selection module}
\newacro{llm}[LLM]{large language model}

\begin{document}
% \title{VLM-SynthDetect: Bridging Vision and Language for Robust Synthetic Image Detection}
\title{Bi-LORA: A Vision-Language Approach for Synthetic Image Detection}

% \title{ AI-Synthesized Image Detection using Large Vision Language Models}

\author{
  Mamadou Keita,
  Wassim Hamidouche,
  Hessen Bougueffa Eutamene,
  Abdenour Hadid,
  Abdelmalik Taleb-Ahmed
  \thanks{Mamadou Keita,  Hessen Bougueffa Eutamene and Abdelmalik Taleb-Ahmed are with Institut d’Electronique de Microélectronique et de Nanotechnologie (IEMN), UMR 8520, Université Polytechnique Hauts de France, Valenciennes, France }
   \thanks{Wassim Hamidouche is with the Technology Innovation Institute, Abu Dhabi, UAE.}   
   \thanks{Abdenour Hadid is with the Sorbonne Center for Artificial Intelligence, Sorbonne University, Abu Dhabi, UAE.}
}



\maketitle

\begin{abstract}

Advancements in deep image synthesis techniques, such as \acp{gan} and \acp{dm}, have ushered in an era of generating highly realistic images. While this technological progress has captured significant interest, it has also raised concerns about the potential difficulty in distinguishing real images from their synthetic counterparts. %The challenge in detecting deepfakes lies in the continuous evolution of generative models. Existing methods predominantly revolve around training classifiers within the image or feature domains, yet achieving robust generalization remains a persistent challenge.%
This paper takes inspiration from the potent convergence capabilities between vision and language, coupled with the zero-shot nature of \acp{vlm}. We introduce an innovative method called Bi-LORA that leverages \acp{vlm}, combined with \ac{lora} tuning techniques, to enhance the precision of synthetic image detection for unseen model-generated images. The pivotal conceptual shift in our methodology revolves around reframing binary classification as an image captioning task, leveraging the distinctive capabilities of cutting-edge \ac{vlm}, notably \ac{blip}2. Rigorous and comprehensive experiments are conducted to validate the effectiveness of our proposed approach, particularly in detecting unseen diffusion-generated images from unknown diffusion-based generative models during training, showcasing robustness to noise, and demonstrating generalization capabilities to \acp{gan}. The obtained results showcase an impressive average accuracy of 93.41\% in synthetic image detection on unseen generation models. The code and models associated with this research can be publicly accessed at \href{https://github.com/Mamadou-Keita/VLM-DETECT}{https://github.com/Mamadou-Keita/VLM-DETECT}
\end{abstract}

\begin{IEEEkeywords}
Deepfake, Text-to-Image Generation, Visual Language Model, Large Language Model, Image Captioning,  Generative Adversarial Nets, Diffusion Models, Low Rank Adaptation.
\end{IEEEkeywords}
\acresetall 
\section{Introduction}
\IEEEPARstart{T}{he} rapid progress in generative models has ushered in a new era of image synthesis, marked by the effortless creation of remarkably realistic images. This technological progress, while undeniably impressive, has also given rise to significant concerns regarding the widespread distribution of fake images. These \ac{ai}-synthesized images, due to their unrivalled visual fidelity, represent a substantial threat to various facets of society. Noteworthy examples, such as the generation of realistic face images through techniques like \acp{gan} and \acp{dm}, have reached a point where they challenge human perception and erode trust in a number of areas, including social media, politics, the military, and cybersecurity.

The pressing issue of discerning between authentic and \ac{ai}-generated images is underscored by the remarkable realism achieved by contemporary generative models, surpassing human perceptual capabilities. As demonstrated in a quantitative study by Lu {\it et al}.~\cite{lu2023seeing}, human observers grapple with distinguishing genuine images from their \ac{ai}-generated counterparts, achieving an accuracy rate as low as 61.3\%. This challenge extends beyond academic research, permeating real-world applications, presenting both opportunities and emerging threats. The advent of text-to-image models allows users to convert textual descriptions into high-quality visual representations, a technological leap with potential malicious applications, including the spread of misinformation and propaganda. In light of these complex challenges, the primary goal is to prioritize the development of efficient detectors capable of reliably identifying fake images generated by these advanced models.

In the realm of image analysis, a myriad of techniques has been developed to detect \ac{ai}-generated images. While \acp{dm} typically utilize \acp{cnn}or \acp{vit} as their foundational architecture, it's crucial to highlight the fundamental disparities in the underlying mechanisms driving image generation in \acp{dm} compared to their predecessors like \acp{vae} and \acp{gan}~\cite{kingma2013auto}. These distinctions in generative processes create a significant gap between detectors designed for earlier generative models and their ability to identify \ac{ai}-generated images generated by diffusion models. To tackle this discrepancy, a seemingly direct approach is to train a binary classifier using \acp{cnn} or transformer-based models to distinguish diffusion-generated images from genuine ones. However, an empirical study conducted by Ricker {\it et al.}~\cite{ricker2022towards} has unveiled a fundamental weakness inherent in this approach. Despite its initial appeal, this simplistic classification framework demonstrates a serious lack of efficiency in generalizing to accurately discriminate new diffusion-generated images that have never been encountered before.

Recent years have witnessed the emergence of a new generation of powerful deep architectures in the field of computer vision. These models, often referred to as foundational models or large \acp{vlm}, have been trained on vast image-text datasets from the vast expanse of the internet~\cite{schuhmann2021laion}. They have made remarkable strides in terms of performance and have assumed a pivotal role in the contemporary computer vision research landscape. Situated at the intersection of \ac{nlp} and \ac{cv}, these models bridge the gap between textual and visual data, ushering in a groundbreaking era of machine understanding. Prominent instances include CLIP~\cite{radford2021learning}, Flamingo~\cite{alayrac2022flamingo}, BLIP2~\cite{li2023blip}, Flava~\cite{singh2022flava}, Llava~\cite{liu2023visual}, SimVLM~\cite{wang2021simvlm}, Visual-BERT~\cite{li2019visualbert}, InstructBLIP~\cite{dai2305instructblip}, and ViTGPT2\footnote{\label{myfootnote}\href{https://ankur3107.github.io/blogs/the-illustrated-image-captioning-using-transformers/}{https://ankur3107.github.io/blogs/the-illustrated-image-captioning-using-transformers/}}. Within the research community, there exists a substantial body of work dedicated to vision-language tasks that require the simultaneous processing of information from visual and linguistic sources to address intricate inquiries. For example, in \ac{vqa}, an image and a corresponding question are provided as input, and the model generates an answer to the query~\cite{antol2015vqa}. Conversely, in image captioning, the model takes an image as input and produces a natural language description of that image~\cite{vinyals2015show}.

% We introduce an innovative
In this paper, we introduce a pioneering approach for synthetic image detection, harnessing state-of-the-art \ac{vlm}.  Traditional methods use complex feature extraction with \acp{cnn}~\cite{o2015introduction} for image classification. In contrast, our approach, named Bi-LORA, reframes the problem as image captioning, leveraging \acp{vlm}' text generation to differentiate between real and synthetic (i.e., \ac{ai}-generated) images. While a similar approach~\cite{chang2023antifakeprompt} adopts an alternative strategy, such as visual question answering, our focus is on generating descriptive captions for images. Specifically, we label authentic images as "true" and synthetic images as "fake", capitalizing on the text generation capabilities of \acp{vlm} (e.g., \acs{blip}2) combined with \ac{lora} tuning strategy for memory efficiency and faster training. Our primary focus is on detecting diffusion-generated images, a challenge distinct from previous generative methods like \acp{gan}. To the best of our knowledge, this work pioneers the concept of treating binary classification as image captioning, harnessing cutting-edge \acp{vlm}. Instead of binary classification, we emphasize the potential of using \acp{vlm} (e.g., \acs{blip}2) to create informative captions indicating class membership. In summary, our work makes the following pivotal contributions:
\begin{itemize}
\item Reconceptualizing binary classification as image captioning: We redefine binary classification as an image captioning task, harnessing the capabilities of \acp{vlm}.
\item Revealing the potential of \acp{vlm} in synthetic image detection: We shed light on the vast potential of \acp{vlm} in the realm of synthetic image detection, showcasing their robust generalization capabilities, even when faced with previously unseen diffusion-generated images.
\item Empirical validation of enhanced detection: Through rigorous and comprehensive experiments, we substantiate the effectiveness of our proposed approach, particularly in the context of detecting diffusion-generated images, robustness to noise and generalization to images generated by \ac{gan}. 
% \hl{ici t as garde icassp, il faut changer par ce qui a ete ajouter sur la robustess au bruit et generalisation GAN}.
\end{itemize}

Our proposed approach, trained only on LSUN Bedroom and \ac{ldm} generated images, far exceeds baseline methods: ResNet50~\cite{he2015deep}, Xception~\cite{Chollet_2017_CVPR}, \ac{deit}~\cite{touvron2021training}, and ViTGPT2. Our outstanding performance and robust generalization stem from the well-aligned vision language representation of pretrained \acp{vlm} and with fewer trainable parameters. Experimental results showcase an impressive average accuracy \( \bf93.41\% \) in synthetic image detection.

The remainder of this paper is organized as follow.  Section \ref{sec:relatedWork} provides a brief review of related works. Section~\ref{sec:VLMsDetect} describes the proposed \ac{vlm}-based approach for synthetic image detection task. Then, the performance of the proposed approach is assessed and analysed in Section~\ref{sec:resultsAnalysis}. Finally, Section~\ref{sec:conclusion} concludes the paper.

% \begin{figure*}[]
%     \centering
%     \includegraphics[width=\linewidth]{figures/HLevel.png}
%     \caption{High level representation of the approach.}
%     \label{fig:HighLevelApproach}
% \end{figure*}

\begin{figure}
    \includegraphics[width=.15\textwidth]{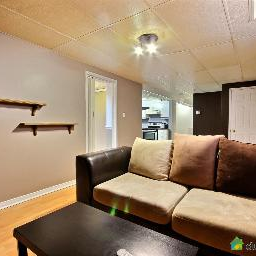}\hfill
    \includegraphics[width=.15\textwidth]{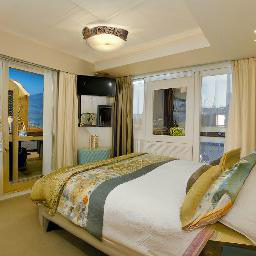}\hfill
    \includegraphics[width=.15\textwidth]{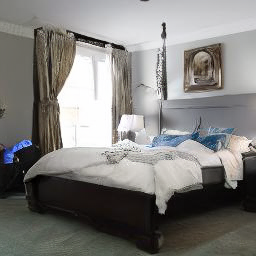}\hfill
    \\[\smallskipamount]
    \includegraphics[width=.15\textwidth]{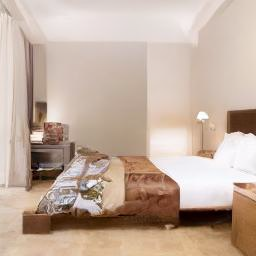}\hfill
    \includegraphics[width=.15\textwidth]{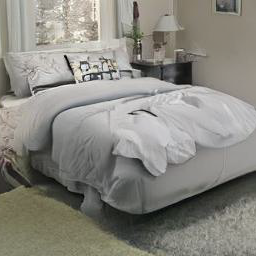}\hfill
    \includegraphics[width=.15\textwidth]{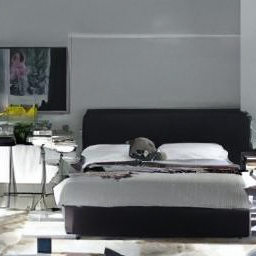}\hfill
    \\[\smallskipamount]
    \includegraphics[width=.15\textwidth]{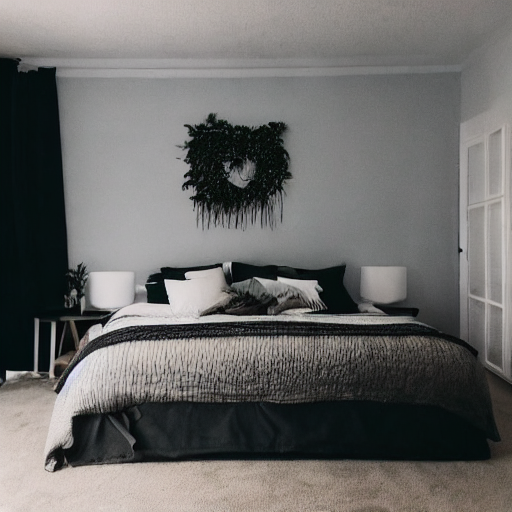}\hfill
    \includegraphics[width=.15\textwidth]{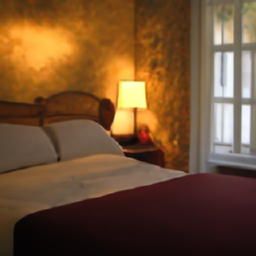}\hfill
    \includegraphics[width=.15\textwidth]{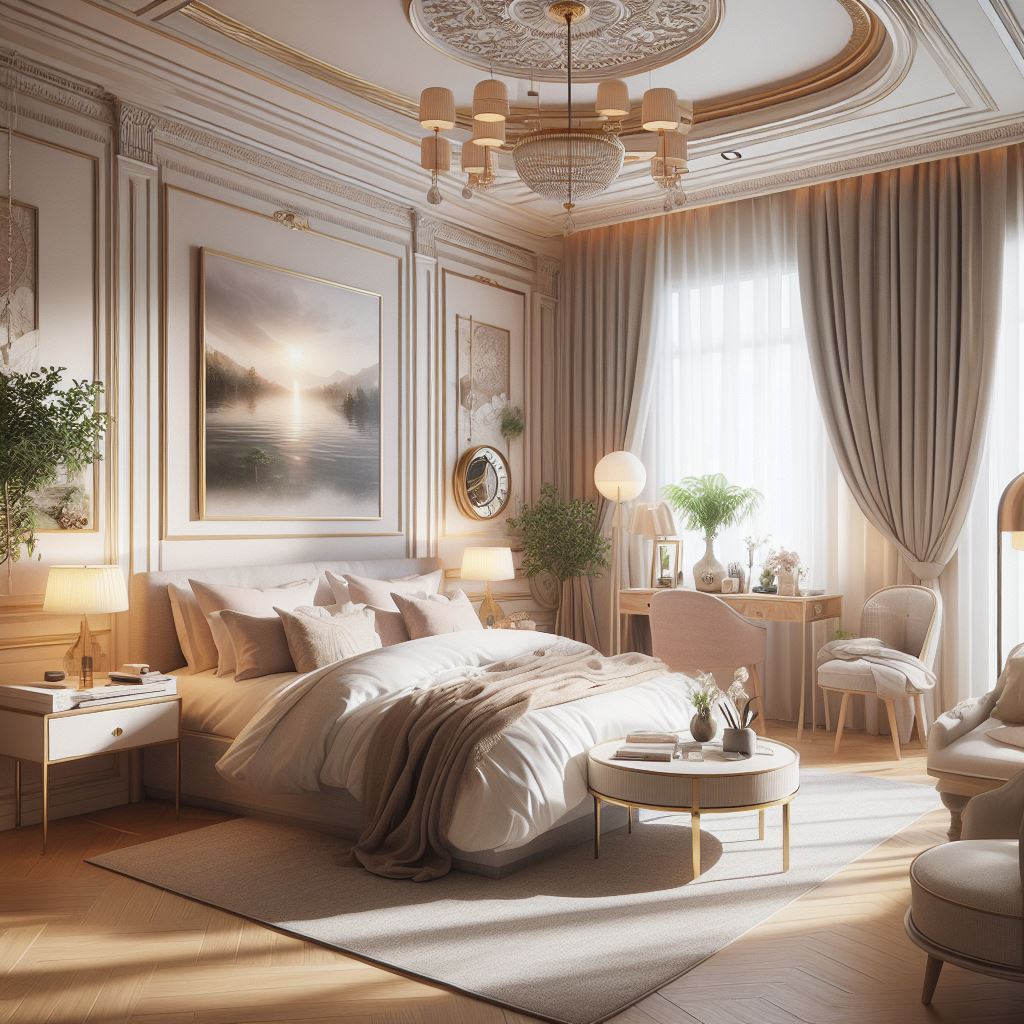}\hfill
    \caption{Examples of real sample and synthetic images generated by various generators in our experiments. From left to righ, Top row: Real~\cite{yu2015lsun}, ADM~\cite{dhariwal2021diffusion}, \acs{ldm}. Middle row: DDPM~\cite{ho2020denoising}, IDDPM~\cite{nichol2021improved}, PNDM~\cite{liu2022pseudo}. Bottom row: Stable Diffusion v1.4~\cite{rombach2022high}, GLIDE~\cite{nichol2021glide}, DALL·E 3~\cite{betker2023improving}.}
    \label{fig:dataset}
\end{figure}
\section{Related Work}
\label{sec:relatedWork}
In this section, we delve into the realm of synthetic image generation, providing an overview of state-of-the-art methods for both image generation and synthetic image detection.
\subsection{Image Generation}
Image generation stands at the intersection of computer vision and machine learning, focusing on the creation of synthetic images through the use of deep learning-based generative models. Prominent among these models are \acp{gan} and, more recently, \acp{dm}. The primary goal of these image generation models is to create visually realistic images that are nearly indistinguishable from real-world images. This field has experienced rapid growth, witnessing numerous advancements in recent years, thereby unlocking a spectrum of applications, including data augmentation, style transfer, and semantic manipulation.

A pioneering method for generating synthetic images was introduced by Goodfellow {\it et al.}~\cite{goodfellow2014generative}. This approach consists of a generator network that generates images and a discriminator network trained to distinguish between real and \ac{ai}-generated images. Since then, this framework named \acp{gan} has been successfully applied to various tasks such as face synthesis~\cite{radford2015unsupervised, karras2017progressive, karras2019style, mokhayeri2020cross, ruiz2020morphgan, liz2024generation}, style transfer~\cite{xu2021drb,chen2018cartoongan, yang2019controllable}, and super-resolution~\cite{zhu2020gan,ledig2017photo,sajjadi2017enhancenet,park2018srfeat,wang2018esrgan}. In \ac{gan}-based image generation, significant progress has been made with the introduction of StyleGAN~\cite{karras2019style}.  This architecture exploits style-based techniques to generate more realistic and compelling images. By incorporating style information into the generating network, StyleGAN gives greater control over the images generated and produces high-quality results. Several variants and extensions of \ac{gan} architectures have been proposed to meet specific challenges and tasks in the field of image generation, such as unpaired data in image generation~\cite{zhu2017unpaired}, allowing control over specific image characteristics~\cite{mirza2014conditional}, and focusing on image generation without labeled data~\cite{radford2015unsupervised}. 
On the other hand, text-to-image generation involves processing a given text description, known as a prompt, to generate a corresponding image. Early endeavors in this domain~\cite{reed2016generative,zhang2017stackgan} utilized \acp{gan} for this purpose. These studies employed the fusion of a prompt's embedding with a latent vector to produce an image that accurately reflects the given description. However, despite the attention garnered by numerous works~\cite{bodla2018semi,lao2019dual,li2019object,souza2020efficient,wang2020text,zhang2021cross}, it is essential to acknowledge that \acp{gan} do not consistently yield satisfactory generation outcomes~\cite{ramesh2021zero,rombach2022high}.

%More recently, a new architecture has been introduced to perform similar tasks, known as diffusion models ~\cite{atwood2016diffusion,nichol2021glide,rombach2022high,saharia2022photorealistic}. These models generate highly realistic images by progressively removing noise from a signal. In text-to-image generation, a given text, known as a prompt, is encoded in a latent vector that guides the process of reducing noisy images into clear ones. Unprecedented results have been achieved with some of these models, e.g. DALL-E ~\cite{ramesh2021zero}, GLIDE ~\cite{nichol2021glide}, Midjourney ~\cite{midjourney}, Imagen ~\cite{saharia2022photorealistic} and Stable Diffusion ~\cite{rombach2022high}. Such models offer unprecedented control over image generation, enabling users to create highly credible images with a high level of detail.

More recently, a groundbreaking architecture known as \acp{dm}~\cite{atwood2016diffusion,nichol2021glide,rombach2022high,saharia2022photorealistic} has emerged, revolutionizing tasks akin to synthetic image generation. Distinguished by their ability to progressively remove noise from a signal, these models stand out for their capacity to generate exceptionally realistic images. In the realm of text-to-image generation, a provided text prompt undergoes encoding into a latent vector, guiding the transformation process from noisy images to clear and intricate representations. Several leading models, such as DALL-E~\cite{ramesh2021zero}, GLIDE~\cite{nichol2021glide}, Midjourney~\cite{midjourney}, Imagen~\cite{saharia2022photorealistic}, and Stable Diffusion~\cite{rombach2022high}, have showcased exceptional results. These models provide unprecedented control over the image generation process, allowing users to craft highly credible images with an impressive level of detail (see Fig. \ref{fig:dataset}).

\subsection{Detection of Synthetic Images}

With the rapid advancement of generative models, the need for efficient methods to detect synthetic images has become evident. While existing detectors have been effective with traditional models like \acp{gan}~\cite{goodfellow2014generative}, recent progress in generative models, particularly diffusion-based architectures and the latest \ac{gan} models, presents major challenges for current detection techniques.

Recent investigations conducted in~\cite{corvi2023detection,ricker2022towards} have studied the efficacy of existing detectors when applied to novel generative models. The outcomes of these studies indicate that current detectors encounter challenges in generalizing their performance to diffusion-based generative architectures and the latest \ac{gan} models. This limitation underscores the necessity for the development of new techniques capable of adeptly detecting images generated by these advanced models. While substantial progress has been achieved in detecting manipulated images produced by traditional models such as \acp{gan}, the specific realm of detecting images generated by diffusion-based models remains relatively unexplored.

% hl{we say diffusion, and latest GAN, then only diffusion ? Yes, as it is said significant progress has been made in detecting GAN generated images and we focus only on diffusion cuz there is few progress done in that domain}.

In their work~\cite{sha2022fake}, the authors employed the ResNet50~\cite{he2016deep} model to discern between real images and those generated by \ac{ai}. They introduced a detection approach that integrates both images and textual prompts, showcasing its efficacy in improving detection accuracy. This method exhibits promising potential for enhancing overall detection capabilities. Similarly, Coccomini {\it et al.}~\cite{coccomini2023detecting} conducted experiments utilizing straightforward deep learning models such as \acp{mlp} and common \acp{cnn} (ResNet50~\cite{he2016deep} and XceptionNet~\cite{Chollet_2017_CVPR}) to differentiate between real and \ac{ai}-generated images. They also implemented a multimodal detector that leverages CLIP as a feature extractor from both images and their corresponding captions, followed by passing the combined features through an \ac{mlp}.
% \hl{how we can get the prompts, we have only images as input, they use coco captioning dataset and flickr30k those dataset have captions. So to generate image they use those captions. Here the prompts is refer to the caption used to generate image. The approach consists of using CLIP to extract features from image and caption, then combine those extracted features and then pass it through a MLP.}. 

\begin{table*}[!ht]
    \centering
    \caption{Architectures and features of state-of-the-art AI-synthesized images detection models}
    \label{tab:stateOfArt}
    \begin{adjustbox}{width=1\linewidth}
        \begin{tabular}{@{}lccccccc@{}}
            \toprule
            \multicolumn{2}{c}{Authors} & \multicolumn{2}{c}{Architecture} & \multicolumn{2}{c}{Dataset} & \multicolumn{2}{c}{Task} \\
            \midrule
            \multicolumn{2}{l}{Sha {\it et al.}$^\clubsuit$ (2022)~\cite{sha2022fake}} & \multicolumn{2}{c}{\acs{cnn}, \acs{clip} + \acs{mlp}}  & \multicolumn{2}{c}{DE-FAKE} & \multicolumn{2}{c}{binary classification} \\
            \midrule
            \multicolumn{2}{l}{Coccomini {\it et al.}$^\clubsuit$ (2023)~\cite{coccomini2023detecting}} & \multicolumn{2}{c}{\acs{cnn}, \acs{clip} + \acs{mlp}} & \multicolumn{2}{c}{Diffusers} & \multicolumn{2}{c}{binary classification} \\
            \midrule
            \multicolumn{2}{l}{Guarnera {\it et al.}$^\clubsuit$ (2023)~\cite{guarnera2023level}} & \multicolumn{2}{c}{\acs{cnn}} & \multicolumn{2}{c}{Level up the DeepFake detection} & \multicolumn{2}{c}{binary classification} \\
            \midrule
            \multicolumn{2}{l}{Amoroso {\it et al.}$^\clubsuit$ (2023)~\cite{amoroso2023parents}} & \multicolumn{2}{c}{\acs{cnn}} & \multicolumn{2}{c}{COCOFake} & \multicolumn{2}{c}{binary classification} \\
            \midrule
            \multicolumn{2}{l}{Wu {\it et al.}$^\clubsuit$ (2023)~\cite{wu2023generalizable}} & \multicolumn{2}{c}{\acs{clip} + \acs{mlp}} & \multicolumn{2}{>{\centering\arraybackslash}m{0.6\linewidth}}{LSUN, Danbooru, ProGAN, \acs{sd}, BigGAN, GauGAN, styleGAN, DALLE, GLIDE, Guided Diffusion, Latent Diffusion, ImageNet, VISION, Artist, DreamBooth, Midjourney, NightCafe, StableAI, YiJian} & \multicolumn{2}{c}{binary classification} \\
            \midrule
            \multicolumn{2}{l}{Xi {\it et al.}$^\clubsuit$ (2023)~\cite{xi2023ai}} & \multicolumn{2}{c}{\acs{cnn}} & \multicolumn{2}{c}{AI-Gen Image} & \multicolumn{2}{c}{binary classification} \\
            \midrule
            \multicolumn{2}{l}{Lorenz {\it et al.}$^\clubsuit$ (2023)~\cite{lorenz2023detecting}} & \multicolumn{2}{c}{\acs{cnn} + \acs{rf}} & \multicolumn{2}{>{\centering\arraybackslash}m{0.6\linewidth}}{CiFAKE, ArtiFact, DIffusionDB, LAION-5B, SAC, \acs{sd}-v2.1, LSUN-Bedroom} & \multicolumn{2}{c}{binary classification} \\
            \midrule
            \multicolumn{2}{l}{Ju {\it et al.}$^\clubsuit$ (2023)~\cite{ju2023glff}} & \multicolumn{2}{c}{\acs{cnn}} & \multicolumn{2}{c}{LSUN, ProGAN, \(DF^3\)} & \multicolumn{2}{c}{binary classification} \\
            \midrule
            \multicolumn{2}{l}{Sinitsa {\it et al.}$^\clubsuit$ (2023)~\cite{sinitsa2023deep}} & \multicolumn{2}{c}{rule-based method} & \multicolumn{2}{>{\centering\arraybackslash}m{0.6\linewidth}}{Laion-5B, \acs{sd} v-1.4, \acs{sd} v-2.1, DALL·E-Mini, GLIDE [20], DALL·E-2, MidJourney, CycleGAN, ProGAN\({}_e\), ProGAN\({}_t\), BigGAN, StyleGAN, StyleGAN2, GauGAN, StarGAN} & \multicolumn{2}{c}{binary classification} \\
            \midrule
            \multicolumn{2}{l}{Guo {\it et al.}$^\clubsuit$ (2023)~\cite{Guo_2023_CVPR}} & \multicolumn{2}{c}{\acs{cnn}} & \multicolumn{2}{c}{HiFi-IFDL} & \multicolumn{2}{c}{binary classification} \\
            \midrule
            \multicolumn{2}{l}{Cozzolino {\it et al.}$^\clubsuit$ (2023)~\cite{cozzolino2023raising}} & \multicolumn{2}{c}{\acs{clip} + \acs{svm}} & \multicolumn{2}{>{\centering\arraybackslash}m{0.6\linewidth}}{ProGAN, StyleGAN2, StyleGAN3, StyleGAN-T, GigaGAN, (Score-SDE, ADM, GLIDE, eDiff-I, Latent and Stable Diffusion, DiT, DeepFloyd-IF, Stable Diffusion XL, DALL·E 2, DALL·E 3, Midjourney v5, Adobe Firefly, LSUN, FFHQ, ImageNet, COCO, LAION, RAISE} & \multicolumn{2}{c}{binary classification} \\
            \midrule
            \multicolumn{2}{l}{Wang {\it et al.}$^\bullet$ (2023)~\cite{wang2023dire}} & \multicolumn{2}{c}{\acs{cnn}} & \multicolumn{2}{c}{DiffusionForensics} & \multicolumn{2}{c}{binary classification} \\ 
            \midrule
            \multicolumn{2}{l}{Ma {\it et al.}$^\bullet$ (2023)~\cite{ma2023exposing}} & \multicolumn{2}{c}{statistical-based approach, \acs{cnn}} & \multicolumn{2}{c}{CIFAR10, TinyImageNet, CelebA} & \multicolumn{2}{c}{binary classification} \\
            \midrule
            \multicolumn{2}{l}{Chang {\it et al.}$^\maltese$ (2023)~\cite{chang2023antifakeprompt}} & \multicolumn{2}{c}{\acl{vlm} (e.g InstructBLIP)} & \multicolumn{2}{>{\centering\arraybackslash}m{0.6\linewidth}}{MS COCO, Flickr30k, SD2, SDXL, DeepFloyd IF, DALLE-2, SGXL, ControlNet, SD2-Inpainting, LaMa, SD2-SuperResolution, LTE} & \multicolumn{2}{c}{visual question answering} \\
            \midrule
            \multicolumn{2}{l}{Bi-LORA$^\maltese$ (Ours)}  & \multicolumn{2}{c}{\acl{vlm} (e.g \acs{blip}2, ViTGPT2)} & \multicolumn{2}{>{\centering\arraybackslash}m{0.6\linewidth}}{LSUN Bed, \acs{adm}, \acs{ldm}, \acs{ddpm}, \acs{iddpm}, \acs{pndm}, SD v-1.4, GLIDE} & \multicolumn{2}{c}{image captioning} \\
            \bottomrule
\end{tabular}
\end{adjustbox}
 \begin{flushleft}
\scriptsize $^\clubsuit$ Deep feature.  $^\bullet$ \Acl{dm}s unique attribute. $^\maltese$ \Acl{vlm}s. 

\end{flushleft}
\end{table*}

Authors in~\cite{wang2023dire} found that \ac{dm}-generated images exhibit features that are more easily reconstructed by pre-trained \acp{dm} than natural images. In order to identify these features, they presented the \ac{dire} utilizing the reconstruction error of images using \ac{ddim} for inversion and reconstruction. \Ac{dire}  mainly focuses on the initial \(x_0\) timestep and might miss information from intermediate steps during diffusion and reverse diffusion. In contrast, \ac{sedid}~\cite{ma2023exposing} was introduced a novel detection method for diffusion-generated images. This technique leverages diffusion patterns' unique properties, focusing on deterministic reverse and denoising computation errors to enhance detection accuracy. By employing the concept of $(t, \delta\text{-error})$ with noise information during each diffusion step, \ac{sedid} method effectively differentiates real images from generated ones. %In addition, ~\cite{guarnera2023level} introduced a multi-level method for synthetic image detection with carefully designed training sets. This approach allows the detector to learn comprehensive features and understand hierarchical attributes. The study applied this hierarchical approach to three tasks: distinguishing real from \ac{ai}-generated images, discerning between \acp{gan} and \acp{dm}, and recognizing \ac{ai}-specific architectures. 
Furthermore, in the work by Guarnera {\it et al.}~\cite{guarnera2023level}, a multi-level method for synthetic image detection was introduced, incorporating meticulously designed training sets. This approach empowers the detector to acquire comprehensive features and comprehend hierarchical attributes. The study applied this hierarchical approach to three tasks: distinguishing real images from those generated by \ac{ai}, discerning between \acp{gan} and \acp{dm}, and recognizing \ac{ai}-specific architectures.

In their work, \cite{amoroso2023parents} investigated the enhancement of stylistic distinction by decoupling semantic and stylistic features in images. The study introduces a contrastive-based disentanglement method to analyze the role of semantics from textual descriptions and low-level perceptual cues in synthetic image detection. However, achieving semantic-style disentanglement in practice proves challenging due to the need for specialized training datasets. Additionally, authors in~\cite{lorenz2023detecting} proposed \acs{multilid}, a novel method for detecting diffusion-generated images. This approach utilizes local intrinsic dimensionality to estimate densities in \ac{cnn} feature spaces, allowing the distinction of models based on their internal feature distribution densities. The process involves using an untrained ResNet18~\cite{he2016deep} to extract low-dimensional features from synthetic images, applying \acs{multilid} to these features, and training a random forest classifier on the \acs{multilid} scores.
%~\cite{amoroso2023parents} investigated how decoupling semantic and stylistic features in images can enhance the distinction in stylistic domain. The study also presents a contrastive-based disentanglement method to analyze the role of the semantics of textual descriptions and low-level perceptual cues in synthetic image detection. However, achieving semantic-style disentanglement in practice is difficult because it requires specialized training datasets. ~\cite{lorenz2023detecting} introduced \acs{multilid}, a novel method for detecting diffusion-generated images. It uses local intrinsic dimensionality to estimate densities in \acs{cnn} feature spaces, distinguishing models based on their internal feature distribution densities. The process involves using an untrained ResNet18~\cite{he2016deep} to extract low-dimensional features from synthetic images, applying \acs{multilid} to these features, and training a random forest classifier on the \acs{multilid} scores. 

In~\cite{xi2023ai}, a novel approach to detect \ac{ai}-generated images is introduced. The method employs a cross-attention-enhanced dual-stream network designed to efficiently capture various features of such images, with a specific emphasis on texture. The network architecture consists of two streams: the residual stream, utilizing an \ac{srm}~\cite{goljan2014rich} residual extraction module to capture residual information, and the content stream, focusing on low-frequency aspects of the images.

Furthermore,~\cite{wu2023generalizable} proposed a distinctive synthetic image detection method named \ac{lasted}. This approach leverages language-guided contrastive learning and a unique formulation of the detection problem to enhance the extraction of highly discriminative representations from limited data. \Ac{lasted} augments training images with carefully designed textual labels, enabling joint image-text contrastive learning for forensic feature extraction. Unlike traditional classification-based approaches, this method treats synthetic image detection as an identification problem, offering a unique and effective perspective to address the challenges in synthetic image detection.
%In ~\cite{xi2023ai}, a new approach to the detection of \ac{ai}-generated images is introduced. The approach uses a cross-attention-enhanced dual-stream network that efficiently captures various features of such images, including texture. The network architecture comprises two streams: the residual stream, which uses an \acs{srm}~\cite{goljan2014rich} residual extraction module to extract residual information, and the content stream, which focuses on low-frequency aspects of the images. Further, ~\cite{wu2023generalizable} introduced a novel synthetic image detection method named \acl{lasted}. This approach utilizes language-guided contrastive learning and a unique formulation of the detection problem to enhance the extraction of highly discriminative representations from limited data. Specifically, \acs{lasted} augments training images with carefully designed textual labels, enabling joint image-text contrastive learning for forensic feature extraction. Unlike traditional classification-based approaches, this method treats the synthetic image detection problem as an identification problem, offering a distinctive and effective approach to address synthetic image detection.

\begin{figure*}[!ht]
    \centering
\includegraphics[width=\linewidth]{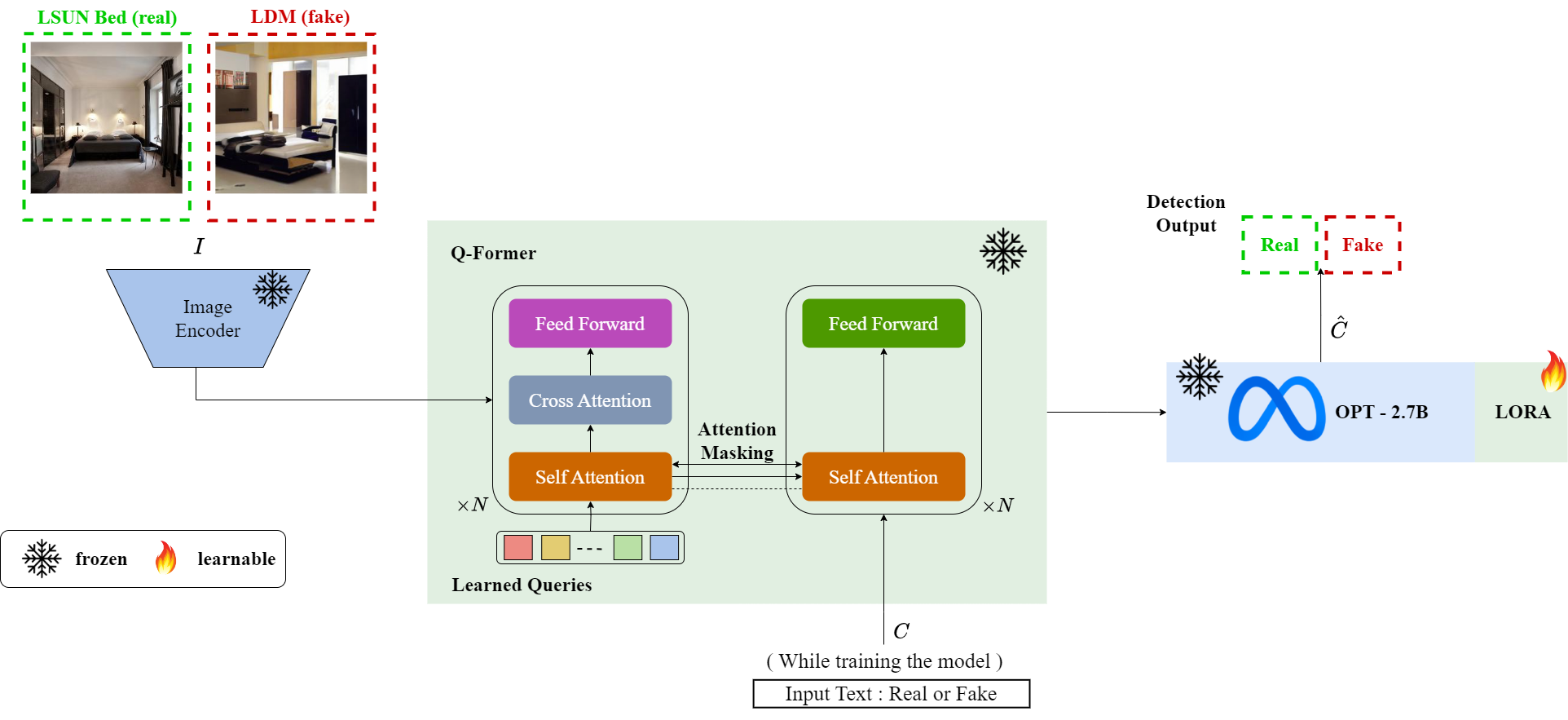}
    \caption{Bi-LORA finetuning for synthetic image detector.}
    \label{fig:architecture}
\end{figure*}

Furthermore, a two-branch global and local feature fusion method was proposed in~\cite{ju2023glff} for the detection of \ac{ai}-synthesized images. This method combines multi-scale global features and informative local features using a ResNet50~\cite{he2016deep} backbone network, known for its efficiency and effectiveness in various face-related tasks. The input image is processed by ResNet50, and an \ac{amsff} module is employed to fuse low-level and high-level features, creating a global representation. This global representation is then fed into a local branch where the \ac{psm} identifies key patches with significant energy and classification relevance. Local features are extracted from these selected patches and integrated with the global representation using an attention-based module for final binary classification.

In contrast to conventional rule-based techniques~\cite{odena2016deconvolution, li2020identification, chandrasegaran2022discovering}, Sinitsa {\it et al.}~\cite{sinitsa2023deep} introduced a novel rule-based method that achieves high detection accuracy, even when trained with a small set of generative images (fewer than 512). This approach leverages the inductive bias of \acp{cnn} to extract distinctive features (fingerprints) from various generators in the training data, enabling the detection of generative images from the same model and its fine-tuned versions.

More recently, {\it Chang et al.}~\cite{chang2023antifakeprompt} drew inspiration from the zero-shot capabilities of \acp{vlm} and proposed an approach that leverages \acp{vlm} such as InstructBLIP. The method incorporates prompt tuning techniques to enhance deepfake detection accuracy over unseen data. The authors reframed the deepfake detection problem as a visual question-answering task, fine-tuning soft prompts for InstructBLIP to distinguish whether a query image is real or fake. On a different note, Cozzolino {\it et al.}~\cite{cozzolino2023raising} introduced a lightweight detection strategy based on \ac{clip} features. These features are employed to design a linear \ac{svm} classifier, and the proposed strategy is evaluated for its performance in challenging scenarios. Finally, Table~\ref{tab:stateOfArt} summarizes state-of-the-art models proposed for the synthetic image detection problem.

%More recently,~\cite{chang2023antifakeprompt} draws inspiration from zero-shot capabilities of \acp{vlm}, and propose an approach harnessing \acp{vlm}, such as InstructBLIP, along with prompt tuning techniques to improve the deepfake detection accuracy over unseen data. They reframe the deepfake detection problem as a visual question answering task, and fine tune soft prompts for InstructBLIP to distinguish whether a query image is real or fake. ~\cite{cozzolino2023raising} introduces a lightweight detection strategy based on \acs{clip} features used to design a linear \acs{svm} classifier and evaluates its performance in challenging scenarios. 

\section{Synthetic Image Detection with VLMs}
\label{sec:VLMsDetect}

In this section, we present our approach for addressing synthetic image detection using large-scale \acp{vlm}. Our approach reframes the binary classification task into an image captioning task, leveraging cutting-edge \acp{vlm} models (e.g., \acs{blip}2). Fig.~\ref{fig:architecture} provides a visual representation of our methodology built around the \ac{blip}2 architecture.

\subsection{Synthetic Image Detection}
\label{sec:SyntheticImageDetection}

Synthetic image detection is generally a binary image classification task. Its main objective is to develop a model \(\mathcal{M}\) that learns a function \(f : \mathbf{I} \rightarrow \mathbf{Y}\) from the training set \(D = \{({I}_i, {y}_i) | 1 \leq i \leq n\}\), where \({I}_i \in \mathbf{I} = \mathbb{R}^{d \times d}\) is an image, and \(y_i \in \mathbf{Y} = \{0, 1\}\) represents the class labels. Here, \(0\) indicates images captured from the real world, and \(1\) indicates generated images. The model's objective is to predict the class \(\hat{y}\) for an input image \({I}\) as follows: 
\begin{equation}
    \hat{y} = f_{\theta}({I}),
\end{equation}
where \(\hat{y}\) is the predicted class label (0 or 1), and \(\theta\) represents the model parameters. 

\subsection{VLM Fine-Tuning for Synthetic Image Detection}
\label{sec:FineTuning}

In the previous subsection, we introduced synthetic image detection from a general perspective and will now proceed to reframe it as an image captioning problem, harnessing the capabilities of \acp{vlm}. Instead of treating synthetic image detection as a traditional binary classification task, we will employ \acp{vlm} to generate descriptive captions for each image. The \acp{vlm} will learn to produce captions that capture the essence of the image, and these captions will serve as indicators of the image's authenticity. Specifically, the model will generate captions that fall into one of two categories: "real" or "fake."

Let's define this fine-tuning process mathematically. We have a \ac{vlm} \(\mathcal{M}\) with parameters $\theta$, which takes an image \(I\) as input and generates a caption \(\hat{C}\). We denote this process as:

\begin{equation}
\hat{C} = \mathcal{M}_{\theta}(I).
\end{equation}

The caption \(\hat{C}\) is a human-like label (textual description) of the image, and it will be used to distinguish the "real" image from the "fake".

To tune the model for this task, we have a dataset \(D = \{(I_i, C_i) | 1 \leq i \leq n\}\), where \(I_i\) is the \(i\)-th image, and \(C_i\) is the ground truth caption whether the image is "real" or "fake". The model's objective is to minimize a suitable loss function \(\mathcal{L}\) over this dataset:

\begin{equation}
\theta^* = \arg\min_\theta \sum_{i=1}^{n} \mathcal{L}(\mathcal{M}_{\theta}(I_i), C_i).
\end{equation}

%Once fine-tuned, the \ac{vlm} will generate descriptive captions to classify images into desired categories, enhancing the synthetic image detection process. 
Once fine-tuned, the \ac{vlm} will be capable of generating captions that help categorizing images into the desired class, providing a more detailed and expressive approach to synthetic image detection.
In the following subsections, we will delve into the details of the fine-tuning process of our Bi-LORA method.
% two \acp{vlm}, ViTGPT2 and \ac{blip}2

\subsection{\acs{lora} Tuning Technique}
\label{sec:loraTuning}

Fine-tuning large pre-trained models is a computational challenge, often involving the adjustment of millions of parameters. This traditional tuning approach, while efficient, requires considerable resources and computational time, creating a bottleneck for the adaptation of these models to specific tasks. \Ac{lora}~\cite{hu2021lora} presents an effective solution to this problem by decomposing the update matrix during fine-tuning, Fig. \ref{fig:lora} (left).

In the traditional fine-tuning method, we modify the weights of a large pre-trained model to adapt it to a new task. This adaptation involves modifying the model's original weight matrix $W$. Changes made to $W$ during fine-tuning are collectively represented by $\Delta W$, so the updated weights can be expressed as $W + \Delta W$. Instead of directly modifying $W$, the \ac{lora} approach seeks to decompose $\Delta W$. Such decomposition is a crucial step in reducing the computational overhead involved in fine-tuning large models.

Intrinsic rank assumption suggests that significant changes in the large pre-trained model can be captured using a lower-dimensional representation. Fundamentally, it asserts that not all elements of \(\Delta W\) are equally important; instead, a smaller subset of these changes can effectively capture the necessary adjustments. Thus, based on this assumption, \ac{lora} proposes to represent \(\Delta W\) as the product of two smaller matrices, $A$ and $B$, with a lower rank. Thus, the updated weight matrix $W'$ becomes:

\begin{equation}
    W' = W + \Delta W  = W + BA.
\end{equation}

Here, $W$ remains fixed (i.e. it is not updated during training). $B$ and $A$ are low-dimensional matrices, and their product $BA$ represents a low-rank approximation of $\Delta W$.

Choosing lower-rank matrices $A$ and $B$ via the \ac{lora} technique results in a notable decrease in trainable parameters, providing advantages in memory efficiency, faster training, adaptability, hardware compatibility, and scalability for large neural networks during fine-tuning.

% \subsection{ViTGPT2}
% \label{sec:vitgpt2}
% ViTGPT2\footnote{\label{myfootnote}\href{https://ankur3107.github.io/blogs/the-illustrated-image-captioning-using-transformers/}{https://ankur3107.github.io/blogs/the-illustrated-image-captioning-using-transformers/}} is a vision encoder-decoder model that belongs to the family of models with a pre-training objective that involves directly fusing visual information into the layers of a language model decoder. As the name suggests, ViTGPT2 adopts the \ac{vit} as the image encoder and the \ac{gpt}-2 model as the language model. For our novel synthetic image detection approach, we fine-tuned ViTGPT-2 using Hugging Face's Seq2SeqTrainer. Initially, we instantiated the model using the VisionEncoderDecoderModel class, which represents an image-to-text model. This model combines the features learned by a transformer-based vision model (the encoder) with the language comprehension capabilities of a pretrained language model (the decoder) available in Hugging Face's utilities. Specifically, we initialized the model with a pretrained \ac{vit} as the encoder and a pretrained language model, GPT-2, as the decoder. Both the encoder and decoder layers are frozen. The VisionEncoderDecoderModel class automatically adds a cross-attention layer to the decoder. This layer is the one trained during fine-tuning.

% \hl{add here what is freeze. All the encoder and decoder are freeze. With VisionEncoderDecoderModel class a cross-attention layer is automallically added to decoder it's that layer who is trained during finetuning}

\subsection{\acs{lora} Tuning on Bi-LORA}
\label{sec:blip2}

\begin{figure}[t]
    \centering
\includegraphics[width=\linewidth]{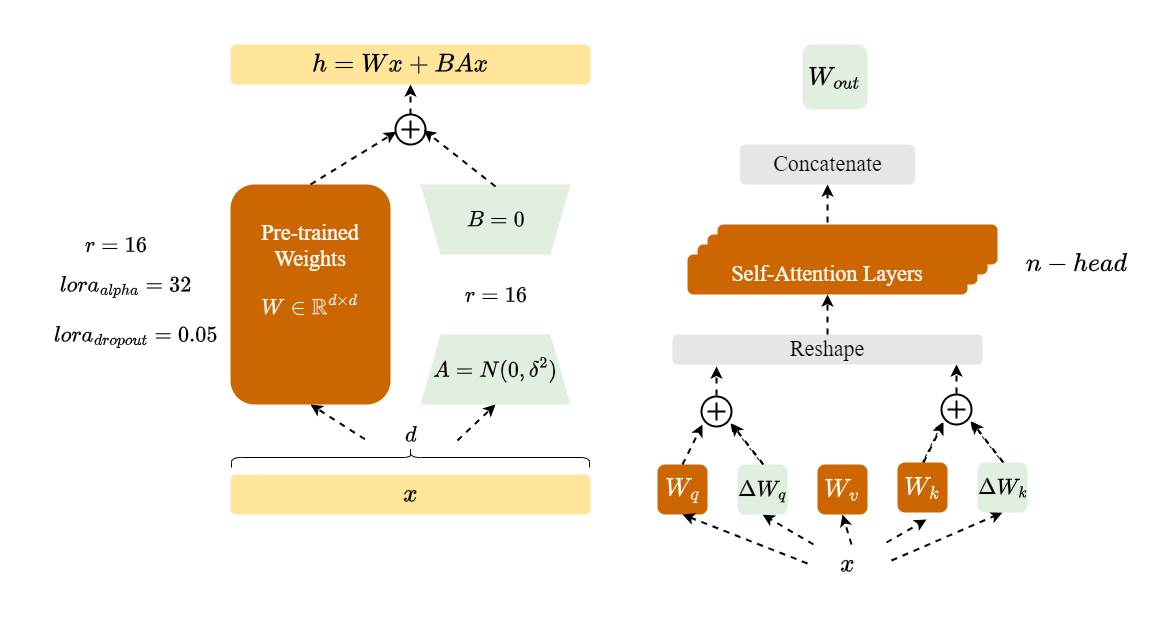}
    \caption{Bi-LORA fine-tuning with \ac{lora}.}
    \label{fig:lora}
\end{figure}

\ac{blip}2~\cite{li2023blip} is a method that improves vision-language alignment using a \ac{q-former} as a bridge. \Ac{q-former} extracts relevant visual information from frozen image encoders and provides it to a frozen language model. The method undergoes two pre-training phases: one for learning visual representations for text and another for training the \ac{q-former} to provide interpretable visual input to the language model. Thus, the smaller matrices are trained to learn task-specific information using supervised learning.
Figure~\ref{fig:lora} (right) provides a visual overview of fine-tuning the \ac{blip}2 model using the \ac{lora} technique. During fine-tuning, we keep the pretrained \ac{blip}2 model weights (\(W\)) frozen and inject a pair of trainable rank decomposition matrices (\(A\) and \(B\)) into each self attention layer, enabling us to optimize and adapt these layers to our new detection task without incurring excessive computational overhead. Thus, the smaller matrices are trained to learn task-specific information using supervised learning. This method is commonly referred to as \ac{lora}~\cite{hu2021lora}, which falls under the re-parameterization category of \ac{peft} methods. Specifically, the update of \ac{blip}2 pre-trained weight matrix $W$ is constrained by a low-rank decomposition $W + \Delta W = W + BA$. Note that both $W$ and $\Delta W = BA$ are multiplied by the same input $x$, and their respective output vectors are summed coordinate-wise. For $h = Wx$, the modified forward pass yields:
\begin{equation}
    h = Wx + \Delta W x = Wx + BAx.
\end{equation}

It is worth highlighting that, among various attention weights, we adapt only two types: \(W_q\) and \(W_k\) of the \ac{llm} decoder module. The first consequence of this technique is that the number of parameters to train is really small, Table \ref{tab:params-table}. However, we run into a second, maybe more important consequence, which is that because we don’t alter the weights of the pretrained model, it doesn’t change its behavior or forget anything it has previously learned, thus we can use it for other tasks if need be.

\begin{table}[!ht]
\caption{Number of parameters for each Bi-LORA's module and the number updated with LoRa}
\label{tab:params-table}
\centering
%\begin{adjustbox}{width=\linewidth}
\begin{tabular}{@{}lcc@{}}
\toprule
Modules              & Trainable     & Freeze        \\ \midrule
Query Tokens         & 0             & 24,576         \\
Image encoder        & 0             & 985,952,256     \\
Qformer              & 0             & 105,137,664     \\
Projection Layer     & 0             & 1,968,640       \\
LLM (OPT 2.7b)         & 5,242,880       & 2,651,596,800    \\
Total              & 5,242,880       & 3,744,679,936    \\ \midrule
\multicolumn{3}{c}{Percentage of trainable parameters: 0.14\%} \\ \bottomrule
\end{tabular}
%\end{adjustbox}
\end{table}

% \begin{table}[!ht]
% \centering
% \caption{Number of parameters for each \acs{blip}2's module and the number updated with LoRa}
% \label{tab:params-table}
% \begin{adjustbox}{width=\linewidth}
% \begin{tabular}{|c@{\hspace{8pt}}|c@{\hspace{8pt}}|c@{\hspace{8pt}}|c@{\hspace{8pt}}|c@{\hspace{8pt}}|c@{\hspace{8pt}}|c@{\hspace{8pt}}|}
% \hline
% & Query Tokens & Image encoder & Qformer & Projection & Language model (OPT 2.7b) & Total \\
% &  &  & & layer & (OPT 2.7b) & Total \\
% \hline
% Trainable params & 0 & 0 & 0 & 0 & 5242880 & 5242880 \\
% \hline
% Freeze params & 24576 & 985952256 & 105137664 & 1968640 & 2651596800 & 3744679936 \\
% \hline
% \multicolumn{7}{|c|}{Trainable params: 5,242,880 | All params: 3,749,922,816 | Trainable\%: 0.13981301102065136} \\
% \hline
% \end{tabular}
% \end{adjustbox}
% \end{table}

\section{Experimental Results}
\label{sec:resultsAnalysis}

This section discusses the empirical results of our proposed synthetic image detection methodology. We present an overview of the experimental setup, delve into cross-generator evaluations, and assess the model's resilience to adversarial conditions.

\subsection{Experimental Setup}
\label{sec:expSetup}
\noindent \textbf{Dataset.} To comprehensively evaluate the performance of our proposed method in synthetic image detection, we employed an existing dataset introduced by Ricker {\it et al.}~\cite{ricker2022towards}. This dataset incorporates real images sourced from the \ac{lsun} bedroom dataset~\cite{yu2015lsun}. We curated a collection of images generated by five distinct diffusion models, all trained on \ac{lsun} bedroom dataset~\cite{yu2015lsun}. Among these models, four subsets of generated images (ADM~\cite{dhariwal2021diffusion}, DDPM~\cite{ho2020denoising}, iDDPM~\cite{nichol2021improved}, and PNDM~\cite{liu2022pseudo}) are generated by unconditional diffusion models. The fifth subset (\ac{ldm}~\cite{rombach2022high}) is generated by text-to-image diffusion model. Further, for generalization evaluation to text-to-image generation, we expanded the dataset by incorporating two additional models: \ac{sd}~\cite{rombach2022high} and GLIDE~\cite{nichol2021glide}. The text prompt used to generate these images was "A photo of a bedroom".

In our experiment, all the subsets consist of 42,000 generated images along with their corresponding real samples from the \ac{lsun} bedroom dataset~\cite{yu2015lsun}. Each subset is divided into 40,000 images for training, 1,000 for validation, and 10,000 for testing purposes. It is worth noting that the real images used for testing is consistent across all the testing subsets. Further details regarding the dataset are provided in the GitHub codebase, in Fig.~\ref{fig:dataset}, we present a selection of representative examples. \\\\
\noindent {\textbf{Evaluation metrics.}} Following the convention of previous detection methods~\cite{ma2023exposing}~\cite{coccomini2023detecting}~\cite{lorenz2023detecting}, we report the accuracy (ACC) and F1 score (F1-Score) in our experiments.  Accuracy (ACC) measures the proportion of correct predictions, with higher scores indicating better performance. While F1-Score balances precision and recall, offering a single measure of a model's accuracy in identifying positive instances while minimizing false results. \\\\ 
{\textbf{Baselines.}} To comprehensively evaluate and compare our proposed approach, we tailored several prominent models, namely ResNet50~\cite{he2015deep}, Xception~\cite{Chollet_2017_CVPR},  \ac{deit}~\cite{touvron2021training}, and ViTGPT2\footref{myfootnote} trained on our dataset, to establish baseline benchmarks. We opted for these models due to their extensive usage and outstanding performance in related tasks. To formulate our baseline models, we fine-tuned these architectures by substituting their final \ac{fc} layers with a novel  \ac{fc} layer featuring a single neuron dedicated to discerning the authenticity of images. These models were initialized with pre-trained weights gleaned from the ImageNet dataset~\cite{deng2009imagenet}, thereby harnessing the knowledge encoded in their learned representations. Employing these custom-tailored baselines enables us to assess the efficacy of our proposed approach when compared with well-established and widely-adopted image classification architectures. \\\\
{\textbf{Implementation details.}} In our experiments, we leveraged the PyTorch deep learning framework on a Windows 11 Pro computer equipped with an NVIDIA RTX A4500 GPU with 16 GB and an Intel(R) i9-12950HX CPU. Our study used baseline models obtained from their publicly available repositories, which we fine-tuned to align with our specific experimental setup. For Bi-LORA model tuning, we opted for the Adam~\cite{kingma2014adam} optimizer with default settings, a learning rate of 5e-5, and training epoch of 20 epochs. During training, we configured the \ac{peft} module, in particular \ac{lora}, with the followings parameters: a \textit{rank}  of 16, \textit{lora\_alpha} set to 32, \textit{lora\_dropout} at 0.05, and a \textit{batch\_size} of 32. Regarding ViTGPT2 model, we followed a procedure outlined by the author, as detailed in provided note\footref{myfootnote}. To ensure a fair comparison, we retrain all baselines on our training dataset, instead of using their published versions directly.

\subsection{Cross-Generator Synthetic Image Detection}
\label{sec:crossValidation}

\begin{table*}[]
\caption{Results of cross-validation on different training and testing subsets using Bi-LORA. We report ACC (\%) / F1-Score (\%) values.}
\label{tab:crossValidation}
\begin{adjustbox}{max width=\textwidth}
\begin{tabular}{@{}cccccccccl@{}}
\toprule
\multirow{2}{*}{Training Subset} & \multicolumn{7}{c}{{Testing Subset}} & \multicolumn{2}{c}{\multirow{2}{*}{\begin{tabular}[c]{@{}c@{}}Average\\ (in \%)\end{tabular}}} \\ \cmidrule(lr){2-8}
 & \acs{ldm}$^\star$ & \acs{adm}$^\oplus$ & \acs{ddpm}$^\oplus$ & \acs{iddpm}$^\oplus$ & \acs{pndm}$^\oplus$ & \acs{sd}$^\star$ & GLIDE$^\star$ & \multicolumn{2}{c}{} \\
 \cmidrule(r){1-1} \cmidrule(l){9-10} 
% \multicolumn{8}{c}{\Acf{blip}} \\
\acs{ldm} & \textbf{99.12 / 99.13} & 85.24 / 82.97 & 98.47 / 98.47 & 97.02 / 96.97 & 99.22 / 99.23 & 77.68 / 71.79 & 97.09 / 97.05 & \multicolumn{2}{c}{\textbf{93.41 / 92.23}                 } \\
\acs{adm} & 95.91 / 95.91 & \textbf{96.51 / 96.57} & 97.24 / 97.31 & 97.35 / 97.42 & 96.16 / 96.21 & 48.32 / 3.35 & 70.99 / 61.90 & \multicolumn{2}{c}{86.07 / 78.38} \\
\acs{ddpm} & 97.94 / 97.94 & 90.49 / 89.76 & \textbf{98.74 / 98.76} & 98.52 / 98.53 & 98.64 / 98.66 & 50.70 / 7.19 & 74.22 / 66.36 & \multicolumn{2}{c}{87.04 / 79.60} \\
\acs{iddpm} & 97.06 / 97.03 & 94.16 / 93.94 & 98.72 / 98.74 & \textbf{98.78 / 98.79} & 98.02 / 98.03 & 49.58 / 2.74 & 62.31 / 41.63 & \multicolumn{2}{c}{85.52 / 75.84} \\
\acs{pndm} & 96.66 / 96.58 & 73.82 / 64.96 & 96.9 / 96.84 & 91.82 / 91.19 & \textbf{99.52 / 99.53} & 85.37 / 83.05 & 97.93 / 97.91 & \multicolumn{2}{c}{91.72 / 90.01} \\
\acs{sd} & 50.14 / 0.58 & 50.00/ 0.00 &  50.00 / 0.00 & 50.00 / 0.00 & 51.74 / 6.73 &  \textbf{100.00 / 100.00} & 99.74 / 99.73  & \multicolumn{2}{c}{ 64.52 / 29.58} \\
GLIDE & 50.72 / 2.88 &  50.00 / 0.00 & 50.03 / 0.14 & 50.0 / 0.02 & 58.38 / 28.71 & 93.56 / 93.12 & \textbf{100.00 / 100.00} & \multicolumn{2}{c}{64.67 / 32.12} \\ \bottomrule
\end{tabular}
\end{adjustbox}
 \begin{flushleft}
\scriptsize $^\star$ Text-To-Image diffusion-based model.  $^\oplus$ Unconditional diffusion-based model.
\end{flushleft}
\end{table*}
Table~\ref{tab:crossValidation} presents the performance metrics of the proposed Bi-LORA model. It was trained on the training set associated with seven generative models and evaluated on their corresponding testing sets. In the table, the diagonal entries reflect the detector's performance when trained and tested on images generated by the same generator. We observe that the detector achieves an accuracy exceeding 96.51\% when both training and testing are conducted on images generated by the same model. Significantly, the subsets \ac{sd} (\ac{sd} V-1.4) and Glide demonstrate precision rates of 100\%. Nevertheless, a notable decline in performance is observed when the model is trained and tested with data generated by different generators. For example, when Bi-LORA is trained on the \ac{adm} and \ac{iddpm} subsets and subsequently tested on \acl{sd} (\ac{sd} V-1.4), the precision diminishes to 48.32\% and 49.58\%, respectively. Conversely, the training subsets of \ac{sd} (\ac{sd} V-1.4) and Glide exhibit poor overall generalization. These two models yield an accuracy of approximately 50\% across all unconditional diffusion-based models and \ac{ldm}, indicating their inefficacy in distinguishing \ac{ai}-generated images from real ones. This inefficiency is further evident in the extremely low F1-Scores of these subsets, suggesting challenges in identifying \ac{ai}-generated images by unconditional diffusion-based models and \ac{ldm}. Furthermore, the experiment highlights \ac{ldm} and \ac{pndm} as the most influential contributors, while \ac{sd} (\ac{sd} V-1.4) and Glide exhibit limitations in performance, especially within our bedroom dataset. Moreover, it is noteworthy that when the Bi-LORA model is trained on an unconditional diffusion-based model, it demonstrates robust generalization to other generative models compared to when trained on a text-to-image model, specifically \ac{sd} (\ac{sd} V-1.4) and Glide.

%For instance, when \acs{blip}2 is trained on the \acs{adm}, \acs{iddpm} subsets and tested on \acl{sd} (\acs{sd} V-1.4), precision drops to 48.32\% and 49.58\%, respectively. In contrast, the \acl{sd} (\acs{sd} V-1.4) and Glide training subsets have poor over all generalization. These two models have accuracy around 50\% over all unconditional diffusion-based model and \acs{ldm}, indicating that they are not effective at distinguishing AI-generated images from real images. This result is reflected in the extremely low F1-Scores of these subsets, suggesting that they struggle to identify AI-generated images by unconditional diffusion-based model and \acs{ldm}. Moreover, through this experiment we can see that \acs{ldm} and \acs{pndm} emerge as the strongest contributors, while \acl{sd} (\acs{sd} V-1.4) and Glide show limits in their performance in our bedroom dataset. Besides, we notice that when the \acs{blip}2 model is trained on an unconditional diffusion-based model, it generalizes well to other generative models than when it is trained on a text-to-image model, more precisely \acl{sd} (\acs{sd} V-1.4) and Glide. It is crucial to note that the diagonal of the table represents scenarios where the training and testing subsets are the same.

\begin{table*}[]
\centering
\caption{Results of different methods trained on \acs{ldm} and evaluated on different testing subsets. We report ACC (\%) / F1-Score (\%).}
\label{tab:TrainLDM}
{\begin{adjustbox}{max width=\linewidth}
\begin{tabular}{@{}lccccccccc@{}}
\toprule
\multirow{2}{*}{Method} & \multicolumn{8}{c}{Testing Subset} & \multirow{2}{*}{\begin{tabular}[c]{@{}c@{}}Average\\ (in \%)\end{tabular}} \\ \cmidrule(lr){2-9}
       & LDM$^\star$    & ADM$^\oplus$    & DDPM$^\oplus$   & IDDPM$^\oplus$  & PNDM$^\oplus$   & SD v1.4$^\star$ & GLIDE$^\star$  \\ \cmidrule(r){1-1} \cmidrule(l){10-10} 
ResNet50 & 99.92 / 99.92      &  72.33 / 61.83     &  75.26 / 67.21    &   88.96 / 87.61    &  77.20 / 70.52     &   75.47 / 67.57    &    73.10 / 63.28   &        &    80.32 / 73.99   \\
Xception & \bf 99.96 / 99.96       &  52.05 / 7.98    &  58.60 / 29.41    &   54.62 / 16.99     &  60.01 / 33.43    &   63.84 / 43.41   &     58.92 / 30.35 &    & 64.00 / 37.36  \\ 
\acs{deit} & 99.83 / 99.83       &  50.40 / 2.01    &   50.18 / 1.17   &  50.14 / 1.01   &    56.25 / 22.54   &   96.02 / 95.86  &    98.15 / 98.11  & &  71.56 / 45.79   \\
ViTGPT2  & 99.40 / 99.40 & 70.84 / 59.21 & 69.60 / 56.72  & 84.08 / 81.20 & 95.40 / 95.22 &  \bf 99.54 / 99.55  &  \bf 99.27 / 99.27 &  & 88.30 / 84.37\\ 
Bi-LORA  & 99.12 / 99.13 & \bf 85.24 / 82.97 & \bf 98.47 / 98.47 & \bf 97.02 / 96.97 & \bf 99.22 / 99.23 &      77.68 / 71.79  &    97.09 / 97.05  &        & \bf 93.41 / 92.23 \\ \bottomrule
\end{tabular}
\end{adjustbox}}
 \begin{flushleft}
\scriptsize $^\star$ Text-To-Image diffusion-based model.  $^\oplus$ Unconditional diffusion-based model.
\end{flushleft}
\end{table*}

\begin{table*}[]
\centering
\caption{ Results of cross-validation on different training and test subsets using different models. Five models trained on seven generators are tested on one generator, and their average accuracy is each data point in the testing subset column. We report ACC (\%) / F1-Score (\%).}
\label{tab:AverageResults}
{\begin{adjustbox}{max width=\linewidth}
\begin{tabular}{@{}lccccccccc@{}}
\toprule
\multirow{2}{*}{Method} & \multicolumn{8}{c}{Testing Subset} & \multirow{2}{*}{\begin{tabular}[c]{@{}c@{}}Average\\ (in \%)\end{tabular}} \\ \cmidrule(lr){2-9}
       & LDM$^\star$    & ADM$^\oplus$    & DDPM$^\oplus$   & IDDPM$^\oplus$  & PNDM$^\oplus$   & SD v1.4$^\star$ & GLIDE$^\star$  \\ \cmidrule(r){1-1} \cmidrule(l){10-10} 
ResNet50 & 75.72 / 54.28 & 69.34 / 43.50 & 61.23 / 25.83 & 76.83 / 55.22 & 61.87 / 27.58 & 77.18 / 60.40 & 82.17 / 68.88 &  & 72.04 / 47.96 \\
Xception & 77.16 / 58.65 & 69.50 / 41.73 & 72.13 / 46.72 & 71.80 / 45.16 & 64.66 / 37.00 & 74.09 / 54.21 & 85.58 / 79.23 &  & 73.56 / 51.82 \\
\acs{deit} & 71.17 / 48.49 & 69.52 / 41.05 & 57.29 / 15.98 & 70.94 / 42.56 & 62.37 / 31.07 & 72.79 / 55.70 & 86.93 / 82.63 &  & 70.14 / 45.35 \\
ViTGPT2  & 77.43 / 60.86 & 67.92 / 44.73 & 74.57 / 56.54 & 78.17 / 61.70 & 84.10 / 69.84 & \bf 76.80 / 60.89 & \bf 89.35 / 87.38 &  & 78.33 / 63.13\\
Bi-LORA  & \bf 83.94 / 70.01 & \bf 77.17 / 61.17 & \bf 84.30 / 70.04 & \bf 83.35 / 68.99 & \bf 85.95 / 75.30 & 72.17 / 51.61 & 86.04 / 80.65 &  & \bf 81.85 / 68.25 \\ \bottomrule
\end{tabular}
\end{adjustbox}}
 \begin{flushleft}
\scriptsize $^\star$ Text-To-Image diffusion-based model.  $^\oplus$ Unconditional diffusion-based model.
\end{flushleft}
\end{table*}

In light of this empirical observation, discerning fake generated images from a specific generator may seem like a straightforward task, achieved by training a detector on a dataset containing both authentic and synthesized images. However, it is crucial to acknowledge that this method may have limitations when confronted with generators of unknown provenance. In practical scenarios, the identity of the generator often remains concealed during the training phase. In the context of our research, we aim to rigorously evaluate the generalization prowess of our Bi-LORA, i.e., its ability to distinguish authentic from synthetic images, regardless of the underlying generator. To this end, we introduce the cross-generator synthetic image detection experiment as an effective means of evaluating the detection capability of the \ac{ai}-generating detector.

%In light of this empirical observation, distinguishing fake generated images by a specific generator appears to be a straightforward task, achieved by training a detector on a dataset containing both genuine and synthesized images. However, it is crucial to acknowledge that this method may exhibit limitations when confronted with generators of unknown provenance. In practical scenarios, the identity of the generator often remains concealed during the training phase. In the context of our research, we aim to rigorously evaluate \acs{vlm}'s generalization prowess, i.e. their ability to distinguish genuine from synthetic images, regardless of the underlying generator. To this end, we introduce the cross-generator synthetic image detection experiment, as an effective means of evaluating the detection capability of the AI-generating detector.

From Table~\ref{tab:crossValidation}, it is evident that the most favorable outcomes are achieved through training on the \ac{ldm}. Consequently, we conduct training of the baseline models on \ac{ldm} and evaluate their performance on test subsets from various generators. The table ~\ref{tab:TrainLDM} also reports the average accuracy across these diverse test subsets.

In all test subsets, the assessed models exhibited varying levels of accuracy in detecting synthetic images. ResNet50~\cite{he2016deep} consistently demonstrated high accuracy, ranging from 72.33\% to 88.96\%, across all tested subsets. Conversely, Xception~\cite{Chollet_2017_CVPR} exhibited relatively lower performance, with accuracy ranging from 52.05\% to 63.84\%. \Ac{deit} showed an intriguing pattern, achieving an accuracy of 96.02\% on the \ac{sd} v1.4 subset, while its accuracy was notably lower on other subsets, suggesting potential challenges in specific synthetic image detection scenarios. Specifically, \ac{deit} performed well in the context of text-to-image generative models (i.e., \ac{ldm}, \ac{sd} v1.4, GLIDE), indicating robust generalization to unseen models from the same subcategory used during training. ViTGPT2 delivered competitive results, with accuracy rates ranging from 69.60\% to an impressive 99.54\%. Notably, Bi-LORA emerged as a standout performer, showcasing outstanding accuracy rates across all subsets, ranging from 77.68\% to 99.22\%. This underscores its robustness in detecting various types of synthetic content.

In Table \ref{tab:AverageResults}, for each model, we have trained seven versions, one for each of the seven generators. We evaluated each version on seven generator test subsets. We calculated the average over seven results for a given model. For example, we respectively trained seven Bi-LORA models using seven generators and calculated the average of their evaluation results on \acs{ldm}, resulting in the highest accuracy of 83.94\%. Our experiment of detecting synthetic images across different generators reveals valuable information about the robustness and limitations of the models evaluated. While some models show solid performance when trained and tested on images from the same generator, challenges emerge in scenarios where identity of the generator is unknown during training. Notably, Bi-LORA stands out as an outstanding performer, demonstrating remarkable accuracy across various generator models.

% \hl{ Our evaluation provides in-depth insights into the capabilities of detectors, aimed at assessing the performance and effectiveness of these architectures in the task of detecting images generated by artificial intelligence. elle n a pas trop de sense cette phrase}

\subsection{Degraded Synthetic Image Detection}
\label{sec:degradedImage}

\begin{figure*}
    \centering
    \begin{subfigure}[b]{0.24\textwidth}
        \includegraphics[width=.8\linewidth]{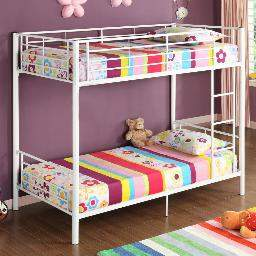}
        \caption{Real - $0|0|0|0|0$ \hspace{8mm} }
    \end{subfigure}
%    \hspace{.1in}
 %   \vspace{.1in}
    \begin{subfigure}[b]{0.24\textwidth}
        \includegraphics[width=.8\linewidth]{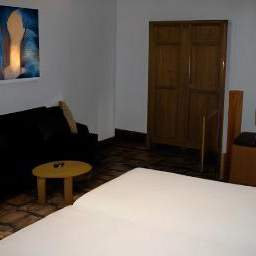}
        \caption{Fake (ADM) - $|0|0|0|0$ \hspace{15mm}}
    \end{subfigure}
   % \hspace{.1in}
   % \vspace{.1in}
    \begin{subfigure}[b]{0.24\textwidth}
        \includegraphics[width=.8\linewidth]{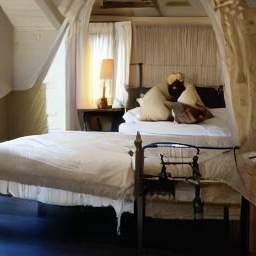}
        \caption{Fake (LDM) - $|0|0|0|1$ \hspace{5mm}}
    \end{subfigure}
   % \hspace{.1in}
   % \vspace{.1in}
    \begin{subfigure}[b]{0.24\textwidth}
        \includegraphics[width=.8\linewidth]{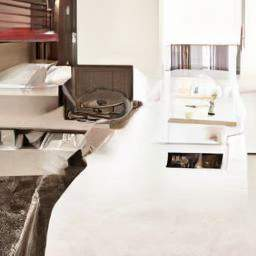}
        \caption{Fake (DDPM) - $|0|0|0|1$ \hspace{5mm}}
    \end{subfigure}
    \begin{subfigure}[b]{0.24\textwidth}
\includegraphics[width=.8\linewidth]{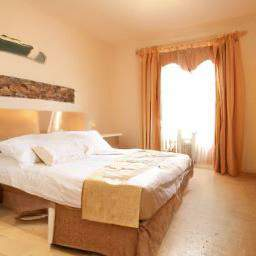}
        \caption{Fake (IDDPM) - $|0|0|0|1$}
    \end{subfigure}
  %  \hspace{.1in}
  %  \vspace{.1in}
    \begin{subfigure}[b]{0.24\textwidth}     \includegraphics[width=.8\linewidth]{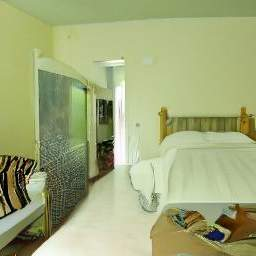}
        \caption{Fake (PNDM) - $|0|0|0|1$}
    \end{subfigure}
%    \hspace{.1in}
%    \vspace{.1in}
    \begin{subfigure}[b]{0.24\textwidth}
        \includegraphics[width=0.8\linewidth]{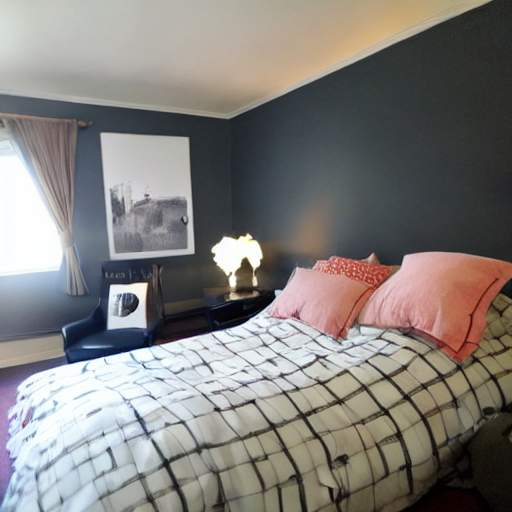}
        \caption{Fake (SD) - $|0|1|1|1$}
    \end{subfigure}
    % \hspace{.1in}
    % \vspace{.1in}
     \begin{subfigure}[b]{0.24\textwidth}     \includegraphics[width=.8\linewidth]{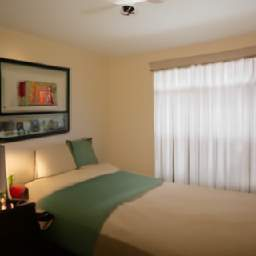}
         \caption{Fake (GLIDE) - $|0|0|0|1$}
     \end{subfigure}  
    \caption{Each sub-figure represents a random compressed (JPEG q=65) image from a testing set, with predicted labels provided below. The 5-digit binary code shows results from ResNet, Xception, \acs{deit}, ViTGPT2, and Bi-LORA models, where '0' means real and '1' means fake. It is important to note that all generated images are considered fake.}
    \label{fig:Visual}
\end{figure*}

In the context of image transmission, images often undergo degradation artifacts due to downscaling to low resolution, compression, and noise interference. Detectors are expected to demonstrate robustness to these distortions. To address this concern, we propose evaluating detectors' performance on degraded images, which more accurately simulate practical conditions, as outlined in Table~\ref{tab:degradedResults}. After training the detectors on the \ac{ldm} subset, we employ various methods to degrade only the test set images. Specifically, we reduce the image resolution to 112 pixels, employing a scaling factor of $2X$. Additionally, we apply JPEG compression with quality settings of 65, introducing compression artifacts. To simulate blurring effects, we apply Gaussian blur with a standard deviation of 3, denoted as $\sigma=3$.

Consequently, the development of a well-thought-out pre-processing approach holds great promise in addressing the challenge of detecting degraded images. Assessing these detectors on degraded images provides valuable insights into their performance under various challenging conditions. This evaluation contributes to a more comprehensive understanding of the detector's capability to handle practical image degradation.

\begin{table}[!ht]
\caption{Models evaluation on degraded images. LR denotes Low Resolution, q denotes quality.}
\centering
\label{tab:degradedResults}
\begin{adjustbox}{width=\linewidth}
\begin{tabular}{@{}cccccccc@{}}
\toprule
\multicolumn{3}{c}{Method}                     & ResNet50      & Xception      & DeiT          & ViTGPT2       & Bi-LORA \\ \midrule
\multirow{31}{*}{\rotatebox{90}{Testing Subset}} & \multirow{3}{*}{\rotatebox{90}{LDM}}     & LR (112) & 82.56 / 81.93 & 61.40 / 67.12  & \bf 98.42 / 98.44 & 93.16 / 93.6  & 95.65 / 95.77 \\ \cmidrule(l){3-8} 
 &                        & JPEG (q=65)        & \bf 95.79 / 95.96 & 50.11 / 0.44  & 50.00 / 0.00  & 54.58 / 16.95 & 90.77 / 89.85                \\ \cmidrule(l){3-8} 
 &                        & Blur ($\sigma$= 3) & 51.36 / 5.57  & 69.56 / 73.61 & 58.88 / 70.77 & 76.65 / 80.83 & \bf 85. 36 / 87.08               \\ \cmidrule(l){2-8} 
 & \multirow{3}{*}{\rotatebox{90}{ADM}}   & LR (112)           & 68.74 / 62.22 & 56.77 / 61.65 & 53.99 / 18.78 & 85.17 / 84.99 & \bf 89.14 / 88.74                \\ \cmidrule(l){3-8} 
 &                        & JPEG (q=65)        & \bf 73.37 / 67.41 & 50.02 / 0.08  & 50.00 / 0.00  & 50.6 / 2.56   & 69.38 / 56.00                 \\ \cmidrule(l){3-8} 
 &                        & Blur ($\sigma$= 3) & 50.38 / 1.76  & 63.20 / 66.24  & 57.20 / 69.21 & 68.58 / 72.37 & \bf 80.53 / 82.07                \\ \cmidrule(l){2-8} 
 & \multirow{3}{*}{\rotatebox{90}{DDPM}}  & LR (112)           & 76.98 / 74.69 & 58.62 / 63.89 & 54.36 / 19.96 & 88.93 / 89.20 & \bf 96.16 / 96.28                \\ \cmidrule(l){3-8} 
 &                        & JPEG (q=65)        & 90.5 / 90.39  & 50.02 / 0.06  & 50.04 / 0.14  & 52.33 / 9.08  & \bf 93.16 / 92.67                \\ \cmidrule(l){3-8} 
 &                        & Blur ($\sigma$= 3) & 51.50 / 6.08  & 65.82 / 69.37 & 56.28 / 68.33 & 74.24 / 78.42 & \bf 85.60 / 87.34                 \\ \cmidrule(l){2-8} 
 & \multirow{3}{*}{\rotatebox{90}{IDDPM}} & LR (112)           & 78.05 / 76.15 & 61.57 / 67.30  & 52.52 / 13.94 & 90.66 / 91.05 & \bf 95.88 / 96.01                \\ \cmidrule(l){3-8} 
 &                        & JPEG (q=65)        & 87.77 / 87.27 & 50.04 / 0.18  & 50.01 / 0.04  & 52.00 / 7.89   & \bf 88.16 / 86.60                 \\ \cmidrule(l){3-8} 
 &                        & Blur ($\sigma$= 3) & 51.10 / 4.55  & 67.58 / 71.41 & 56.74 / 68.78 & 73.91 / 78.08 & \bf 85.37 / 87.07                \\ \cmidrule(l){2-8} 
 & \multirow{3}{*}{\rotatebox{90}{PNDM}}  & LR (112)           & 78.92 / 77.32 & 57.03 / 61.96 & 68.06 / 54.84 & 92.76 / 93.19 & \bf 96.27 / 96.39                \\ \cmidrule(l){3-8} 
 &                        & JPEG (q=65)        & 89.99 / 89.82 & 50.02 / 0.10   & 50.00 / 0.00  & 53.44 / 13.05 & \bf 97.0 / 96.91                 \\ \cmidrule(l){3-8} 
 &                        & Blur ($\sigma$= 3) & 51.15 / 4.78  & 69.4 / 73.43  & 58.30 / 70.24  & 76.28 / 80.46 & \bf 85.76 / 87.49                \\ \cmidrule(l){2-8} 
                                 & \multirow{3}{*}{\rotatebox{90}{SD v1.4}} & LR (112) & 72.80 / 68.66 & 61.68 / 67.42 & \bf 93.25 / 92.96 & 93.18 / 93.61 & 88.10 / 87.49  \\ \cmidrule(l){3-8} 
 &                        & JPEG (q=65)        & 62.72 / 47.53 & 74.40 / 65.59  & \bf 95.72 / 95.53 & 85.06 / 82.45 & 86.26 / 84.10                \\ \cmidrule(l){3-8} 
 &                        & Blur ($\sigma$= 3) & 51.94 / 7.72  & \bf 77.09 / 81.36 & 54.12 / 66.24 & 76.87 / 81.04 & 71.14 / 70.87                \\ \cmidrule(l){2-8} 
 & \multirow{3}{*}{\rotatebox{90}{GLIDE}} & LR (112)           & 79.36 / 77.89 & 66.84 / 73.00  & \bf 96.88 / 96.87 & 93.06 / 93.49 & 93.98 / 94.04                \\ \cmidrule(l){3-8} 
 &                        & JPEG (q=65)        & 89.09 / 88.80  & 50.90 / 3.52   & 50.00 / 0.00  & 57.03 / 24.79 & \bf 98.39 / 98.37                \\ \cmidrule(l){3-8} 
 &                        & Blur ($\sigma$= 3) & 50.94 / 3.95  & 69.57 / 73.62 & 59.01 / 70.89 & 72.31 / 76.42 & \bf 83.06 / 84.76                \\ \bottomrule
\end{tabular}
\end{adjustbox}
\end{table}

Table \ref{tab:degradedResults} presents the evaluation results for different models on degraded images. ResNet50 demonstrates varying performance across different degradation methods, notably excelling under JPEG compression, due to being trained on data that has been compressed with settings ranging from 60 to 100. Conversely, Bi-LORA exhibits robust performance across all degradation methods, highlighting its effectiveness in handling challenges associated with detecting degraded synthetic images 
% \hl{highligh in bold and more analysis; ;ore degradation causes by bur, etc}. 
These results underscore the significance of evaluating detectors under challenging conditions to assess their practical suitability in real-world scenarios and emphasize the necessity for models that can reliably perform in the presence of image degradation.

%In Table \ref{tab:degradedResults}, the evaluation results for different models on degraded images are presented. ResNet50 exhibits varying performance across different degradation methods, performing notably well under JPEG compression due to its training data augmentation with settings ranging from 60 to 100. In contrast, \acs{blip}2 demonstrates robust performance across all degradation methods, showcasing its effectiveness in handling challenges related to detecting degraded synthetic images. These findings underscore the importance of assessing detectors under challenging conditions to gauge their practical suitability in real-world scenarios and emphasize the need for models that can perform reliably in the presence of image degradation.
\begin{table*}[!ht]
\centering
\caption{Evaluation of seven trained Bi-LORA models on five \acs{gan}-based testing subsets. We report ACC (\%) / F1-Score (\%).}
\label{tab:GANGenResults}
%\begin{adjustbox}{width=\linewidth}
\begin{tabular}{@{}ccccccc@{}}
\toprule
\multirow{2}{*}{Training Subset} &
  \multicolumn{5}{c}{Testing Subset} &
  \multirow{2}{*}{\begin{tabular}[c]{@{}c@{}}Average\\ (in \%)\end{tabular}}  \\ \cmidrule(lr){2-6}
      & StyleGAN      & Diff-STyleGAN2 & Diff-ProjectedGAN & ProGAN        & ProjectedGAN  &               \\ \cmidrule(r){1-1} \cmidrule(l){7-7} 
LDM   & 99.09 / 99.09 & 99.10 / 99.10  & 99.08 / 99.08     & \bf 99.26 / 99.26 & 99.23 / 99.23 & 99.15 / 99.15 \\
ADM   & 85.92 / 84.54 & 90.93 / 90.56  & 96.56 / 96.74     & \bf 97.30 / 97.37 & 96.52 / 96.58 & 93.47 / 93.16 \\
DDPM  & 93.80 / 93.56 & 96.24 / 96.19  & 98.31 / 98.32     & \bf 98.80 / 98.81 & 98.43 / 98.45 & 97.12 / 97.07 \\
IDDPM & 89.67 / 88.76 & 93.64 / 93.37  & 97.78 / 97.78     & \bf 98.74 / 98.75 & 97.70 / 97.70 & 95.51 / 95.27 \\
PNDM  & 99.15 / 99.15 & 99.44 / 99.45  & \bf 98.60 / 98.59     & 99.56 / 99.57 & 99.13 / 99.13 & \bf 99.18 / 99.18 \\
SD    & 66.06 / 48.62 & \bf 76.40 / 69.11  & 51.12 / 4.38      & 63.91 / 43.53 & 51.43 / 5.56  & 61.78 / 34.24 \\
GLIDE & 65.36 / 47.02 & 83.45 / 80.17  & 57.42 / 25.86     & \bf 84.42 / 81.54 & 59.81 / 32.82 & 70.09 / 53.48 \\
Average (in \%) &
  85.59 / 80.11 &
  91.31 / 89.71 &
  85.55 / 74.39 &
  \bf 91.71 / 88.40 &
  86.04 / 75.64 &
  88.04 / 81.65 \\ \bottomrule
\end{tabular}
%\end{adjustbox}
\end{table*}

Fig. \ref{fig:Visual} presents visual examples of randomly selected compressed (JPEG q=65) images from testing sets, with predicted labels below each image. The visual examples highlight the challenges posed by image degradation on detection models. In some cases, the models exhibit accurate predictions even in the presence of degradation, while in others, degradation introduces ambiguity, leading to misclassifications. Once again, Bi-LORA consistently demonstrates robust performance across various degradation methods, showcasing its effectiveness in handling challenges related to detecting degraded synthetic images. These visual examples complement the quantitative results presented in Table \ref{tab:degradedResults}, providing a more intuitive understanding of how different models respond to degraded images. The variability in performance across models and degradation methods underscores the importance of comprehensive evaluations under realistic conditions to assess the practical suitability of image detection models.

\subsection{Generalization to \acs{gan} models}
\label{sec:generalizationToGAN}

In this section, our aim is to investigate whether our Bi-LORA approach, primarily tuned for detecting images generated by diffusion-based models, can effectively generalize to \ac{gan}-generated images without the need for additional fine-tuning on \ac{gan}-specific data. One of the key objectives in assessing the generalization of our Bi-LORA approach to \ac{gan}-generated images is to explore the transferability of learned features. Our Bi-LORA approach has acquired knowledge about common features and artifacts present in \ac{ai}-generated images, particularly those created using diffusion-based models. We hypothesize that some of these features may be transferable and applicable to \ac{gan}-generated images. To conduct this analysis, we curated a dataset consisting of various models based on \ac{gan}, including Diff-StyleGAN2~\cite{wang2022diffusion}, Diff-ProjectedGAN~\cite{wang2022diffusion}, ProGAN~\cite{karras2017progressive}, ProjectedGAN~\cite{sauer2021projected}, and StyleGAN~\cite{karras2019style}. Each of these subsets comprises 10,000 generated images along with 10,000 real samples. Our approach involves using this dataset to evaluate the performance of our fine-tuned Bi-LORA approach in distinguishing images generated by \ac{gan} from real images.

In Table~\ref{tab:GANGenResults}, we present the performance evaluation of seven trained Bi-LORA models across five testing subsets based on \acp{gan}. Notably, models such as \ac{ldm}, \ac{adm}, \ac{ddpm}, \ac{iddpm}, and \ac{pndm} consistently exhibit high accuracy and F1-scores. This suggests that our Bi-LORA approach trained on diffusion-based models generalize effectively to images generated by \acp{gan}. The remarkable consistency in performance across different \ac{gan} models indicates that Bi-LORA have successfully learned features and patterns that are transferable between diffusion-based and \ac{gan}-based generative models.

However, StyleGAN presents a challenge, particularly for Bi-LORA model trained on  \acs{sd} and GLIDE generators, where both accuracy and F1-score are considerably lower compared to other \ac{gan} models. This discrepancy may be attributed to the unique architecture and characteristics of StyleGAN, requiring specific features that were not as prominent in diffusion-based models. Bi-LORA trained on different generators exhibits varied performance on \ac{gan} models. For instance, \ac{pndm} consistently performs well across all \ac{gan} subsets, while \ac{sd} and GLIDE show lower performance, indicating differences in feature importance and generalization capabilities among these models.

The findings suggest that our Bi-LORA approach trained on diffusion-based models can generalize to \ac{gan}-generated images to a considerable extent. This has implications for applications where detecting \ac{ai}-generated content is crucial, such as content moderation and deepfake detection.

%In Table~\ref{tab:GANGenResults}, we showcase the performance evaluation of seven trained \acs{blip}2 models across five testing subsets based on \acp{gan}. Notably, models such as LDM, ADM, DDPM, IDDPM and PNDM consistently display high accuracy and f1-scores. This indicates that \acs{vlm} trained on diffusion-based models generalize effectively to images generated by \acs{gan}. The notable consistency of performance between different \ac{gan} models suggests that \acs{vlm} have successfully learned features and patterns transferable between diffusion-based and \ac{gan}-based generative models. However, StyleGAN presents a challenge, particularly  for \acs{sd} and GLIDE models, where both accuracy and F1-Score are considerably lower compared to other \ac{gan} models. This discrepancy may be attributed to the unique architecture and characteristics of StyleGAN, requiring specific features that were not as prominent in diffusion-based models. Different \acp{vlm} exhibit varied performance on \ac{gan} models. For instance, PNDM consistently performs well across all \acs{gan} subsets, while \acs{sd} and GLIDE show lower performance, indicating differences in feature importance and generalization capabilities among these models.

%The findings suggest that \acp{vlm} trained on diffusion-based models can generalize to \acs{gan}-generated images to a considerable extent. This has implications for applications where detecting \ac{ai}-generated content is crucial, such as content moderation and deepfake detection.

\subsection{Comparative Analysis of Synthetic Image Detection Models}
\label{sec:comparaison}

In this section, we perform a comparative analysis of Bi-LORA against four recent synthetic image detection models : \textbf{AntifakePrompt} (Chang {\it et al.}, 2023)~\cite{chang2023antifakeprompt}, \textbf{DE-FAKE} (Sha {\it et al., 2022})~\cite{sha2022fake}, \textbf{UniversalFakeDetect} (Ojha {\it et al., 2023})~\cite{ojha2023universal}, \textbf{CNNDetection} (Wang {\it et al., 2020})~\cite{wang2020cnn}. For AntifakePrompt, we use checkpoints of detector trained on MS COCO vs. SD2 and MS COCO vs. SD2+LaMa. For DE-FAKE, we employ the checkpoint of the hybrid detector, which considers both the image and the corresponding prompts in inference time. For CNNDetection and UniversalFakeDetect, we use the detectors checkpoint trained on a dataset containing images that may have undergone Gaussian blurring and JPEG augmentation, each with a probability of 10\%. 

\begin{table*}[]
\caption{Comparaison with SOTA models}
\label{tab:sota-comparaison}
\begin{adjustbox}{width=\linewidth}
\begin{tabular}{|c|c|
>{\columncolor[HTML]{CCFFCC}}c |
>{\columncolor[HTML]{feb3b1}}c |
>{\columncolor[HTML]{feb3b1}}c |
>{\columncolor[HTML]{feb3b1}}c |
>{\columncolor[HTML]{feb3b1}}c |
>{\columncolor[HTML]{feb3b1}}c |
>{\columncolor[HTML]{feb3b1}}c |
>{\columncolor[HTML]{feb3b1}}c |}
\hline
Methods                          & Training Set         & LSUN Bed & ADM   & DDPM  & IDDPM & PNDM  & LDM   & SD v1.4 & GLIDE \\ \hline
                                 & MS COCO vs. SD2      &     92.86     &    29.76   &   57.05    &   47.99    &  76.79     &  57.21     &    5.80     &     98.86   \\ \cline{2-10} 
\multirow{-2}{*}{AntifakePrompt} & MS COCO vs. SD2+LaMa &     65.58     &  \bf 73.18    &    87.59   &   82.33    &   92.31    &   88.56    &     6.80    &  \bf 99.62    \\ \hline
 DE-FAKE &  MS COCO vs. SD2 &  27.87  &  42.70     &   55.95    &   
 31.69  &  44.24    &   53.43   &     \bf  80.95      &   49.90      \\ \hline
UniversalFakeDetect  & LSUN vs. ProGAN      & \bf 
 99.80    & 06.80 & 36.10 & 24.70 & 47.50 & 34.10 & 21.70   & 12.00 \\ \hline
CNNDetection &  LSUN vs. ProGAN &   \bf   99.80     &   00.90   &     01.67   &    01.46  &   01.46     &   01.30    &     02.62    &   07.92    \\ \hline
% DIRE                             &                      &          &       &       &       &       &       &         &       \\ \hline
% FusingGlobalandLocal             &                      &          &       &       &       &       &       &         &       \\ \hline
% DMimageDetection                 &                      &          &       &       &       &       &       &         &       \\ \hline
% Diffusers                        &                      &          &       &       &       &       &       &         &       \\ \hline

% LGrad                            &                      &          &       &       &       &       &       &         &       \\ \hline

% AutoGAN                          &                      &          &       &       &       &       &       &         &       \\ \hline
% F3Net                            &                      &          &       &       &       &       &       &         &       \\ \hline

Bi-LORA (Ours)                          & LSUN Bed vs. LDM &     98.55     &    71.93    &  \bf    98.39  &  \bf  95.49    &  \bf  99.89    & \bf   99.70    &      56.81   &   95.63  \\ \hline

\end{tabular}
\end{adjustbox}
\end{table*}

\begin{table*}[]
\caption{General Image Datasets}
\label{tab:datasets}
\begin{adjustbox}{width=\linewidth}
\begin{tabular}{ccc
>{\columncolor[HTML]{CCFFCC}}c 
>{\columncolor[HTML]{feb3b1}}c c}
\hline
Dataset       & \#Real     & \#Generated & Source of  Real Image & Generation Method                   & Year \\ \hline
Synthbuster   & -          & 9,000       & Raise-1k              & DALL-E2\&3, Midjourney, SDMs, GLIDE, Firefly & 2023 \\
GenImage      & 1, 331,167 & 1,350,000   & ImagNet               & SDMs, Midjourney, BigGAN            & 2023 \\
CIFAKE        & 60,000     & 60,000      & CIFAR-10              & SD v1.4                             & 2023 \\
AutoSplice    & 2,273      & 3,621       & Visual News           & DALL-E 2                            & 2023 \\
DiffusionDB   & 3,300,000  & 16,000,000  & DiscordChatExporter   & SD                                  & 2023 \\
ArtiFact      & 964,989    & 1,531,749   & COCO, FFHQ, LSUN      & SDMs, DDPM, LDM, CIPS               & 2023 \\
HiFi-IFDL     & \texttildelow600,000   & 1,300,000   & FFHQ, COCO, LSUN      & DDPM, GLIDE, LDM, GANs              & 2023 \\
DiffForensics & 232,000    & 232,000     & LSUN, ImageNet        & LDM, DDPM, VQDM, ADM                & 2023 \\
CocoGlide     & 512        & 512         & COCO                  & GLIDE                               & 2023 \\
LSUNDB &
  420,000 &
  510,000 &
  LSUN &
  \begin{tabular}[c]{@{}c@{}}LDM, ADM, DDPM, IDDPM, PNDM, \\ StyleGAN, ProGAN, Diff-ProjectedGAN,\\  ProjectedGAN, Diff-StyleGAN2\end{tabular} &
  2023 \\
UniFake       & 8,000      & 8,000       & LAIION-400M           & LDM, GLIDE                          & 2023 \\
REGM          & -          & 116,000     & CelebA, LSUN          & 116 publicly available GMs          & 2023 \\
DMimage       & 200,000    & 200,000     & COOC, LSUN            & LDM                                 & 2022 \\
AIGCD         & 360,000    & 508,500     & LSUN, COCO, FFHQ      & SDMs, GANs, ADM, DALL·E 2, GLIDE    & 2023 \\
DIF           & 84,300     & 84,300      & LAION-5B              & SDMs, DALL·E 2, GLIDE, GANs         & 2023 \\
Fake2M        & -          & 2,300,000   & CC3M                  & SD-V1.5, IF, StyleGAN3              & 2023 \\ \hline
\end{tabular}
\end{adjustbox}
\end{table*}

Table~\ref{tab:sota-comparaison} compares the performance of four state-of-the-art (SOTA) models on the task of synthetic image detection. It shows the average accuracy of each model on various test datasets (ADM,DDPM etc.) when trained on different real vs. fake image set combinations (e.g., MS COCO vs. SD2+LaMa). The performance of all models varies depending on the training data used. For example, UniversalFakeDetect and CNNDetection perform exceptionally well on the LSUN Bed test set, but perform poorly on other testing sets. In addition, some test data sets seem to be more challenging than others. For instance, AntifakePrompt, Bi-LORA, CNNDetection and UniversalFakeDetect struggle on the SD v1.4 dataset, suggesting that these models may have difficulty detecting certain types of synthetic images.

AntifakePrompt achieves the highest accuracy on the GLIDE and ADM test set. However, performance drops significantly on the datasets generated by the other models when trained on MS COCO vs. SD2. In contrast, it generalizes quite well on the test sets when trained on the MS COCO vs. SD2+LaMa combination.

Regarding the DE-FAKE detector, trained on images generated by SD and MS COCO, it achieves the highest accuracy on the SD v1.4 test set, however it suffers from drops in accuracy when applied to images generated by unseen generative models. Although the model considers corresponding prompts, potentially helping to detect unusual synthetic scenarios, its effectiveness is limited because most of the training data uses natural prompts from the MS COCO dataset. This biases the model towards identifying false images, making it less effective against fakes generated with different prompts.

UniversalFakeDetect and CNNDetection, both trained on the LSUN vs ProGAN dataset, show outstanding performance (near-perfect accuracy) on LSUN Bed. Meanwhile, their performance suffers significantly on all other datasets, highlighting a major weakness in terms of generalizability and dependence on a particular training data configuration.

Bi-LORA achieves good accuracy on the majority of test sets (Lsun Bed, DDPM, IDDPM, PNDM, LDM) except SD v1.4, suggesting that Bi-LORA is effective in generalizing to different types of synthetic images.

\subsection{Evaluation on Diverse Datasets}
\label{sec:OtherDataset}

The performance of deep learning models in real-world applications depends not only on their effectiveness in the domains on which they were initially trained, but also on their ability to generalize and adapt to diverse and new data sources. In the context of image authenticity assessment, where it is crucial to distinguish synthetic from real-world images, evaluation on diverse datasets becomes an essential aspect of model validation. In this section, we dive into a comprehensive exploration of the generalizability and robustness of our proposed approach.

Our vision language model, preliminarily fine-tuned on the LSUN bedroom dataset, seems initially specialized in a specific task: distinguishing synthetic bedroom scenes from real ones. While this specialization has yielded promising results in its niche, the wider landscape of visual data extends far beyond the confines of bedrooms, Table. \ref{tab:datasets} . In real-world applications, the need to detect image authenticity often arises in scenarios where images encompass a myriad of content, with diverse objects, scenes, and contexts. It is therefore imperative to evaluate the performance of our model when applied to more general datasets.

\begin{table*}[!ht]
\caption{Performance comparison of Bi-LORA trained on a general dataset vs. AntifakePrompt. Experiments are conducted on 2 real and 14 fake datasets, including 3 attacked ones.}
\label{tab:TrainedOnCvprDataResults}
\begin{adjustbox}{width=\linewidth}

\begin{tabular}{@{}ccccccccc@{}}
\toprule
Methods & Training Set & No. of param & MS COCO & Flickr & SD2 & SDXL & IF & DALLE-2 \\ \midrule
AntifakePrompt & MS COCO vs. SD2 & 4.86K & 95.37 & 91.00 & 97.83 & 97.27 & 89.73 & 99.57 \\
 & MS COCO vs. SD2+LaMa & 4.86K & 90.83 & 81.04 & 97.10 & 97.10 & 88.37 & 99.07 \\
Bi-LORA & MS COCO vs. SD2 & 5.24M & \bf92.10 & \bf90.33 & 99.40 & 98.30 & 90.60 &  99.23\\
 &MS COCO vs. SD2+LaMa & 5.24M & 80.30 & 72.53 & \bf99.47 & \bf99.37 & \bf95.03 & \bf99.67 \\ \bottomrule

\toprule
\multirow{2}{*}{Methods} & \multirow{2}{*}{Training Set} & \multirow{2}{*}{No. of param} & \multirow{2}{*}{SGXL} & \multirow{2}{*}{ControlNet} & \multicolumn{2}{c}{Inpainting} & \multicolumn{2}{c}{Super Res.} \\ \cmidrule(l){6-9} 
 &  &  &  &  & LaMa & SD2 & LTE & SD2 \\ \cmidrule(r){1-5}
\multirow{2}{*}{AntifakePrompt} & MS COCO vs. SD2 & 4.86K & \bf99.97 & 91.47 & 39.03 & 85.20 & 99.90 & 99.93 \\
 & MS COCO vs. SD2+LaMa & 4.86K & 99.93 & 93.27 & 58.53 & 90.70 & \bf100.00 & 99.97 \\
 
 \multirow{2}{*}{Bi-LORA} & MS COCO vs. SD2 & 5.24M &  96.80 & 96.87 & 46.47 & 92.30 & 98.33 & 99.73 \\
 & MS COCO vs. SD2+LaMa & 5.24M & 98.50 & \bf98.90 & \bf80.07 & \bf94.13 & 99.67 & \bf 100.00 \\ \bottomrule

\toprule
\multirow{2}{*}{Methods} & \multirow{2}{*}{Training Set} & \multirow{2}{*}{No. of param} & \multirow{2}{*}{\begin{tabular}[c]{@{}c@{}}Deeper-\\ Forensics\end{tabular}} & \multirow{2}{*}{Adver.} & \multicolumn{2}{c}{Attack} & \multicolumn{2}{c}{\multirow{2}{*}{Average}} \\ \cmidrule(lr){6-7}
 &  &  &  &  & Backdoor & Data Poisoning & \multicolumn{2}{c}{} \\ \cmidrule(r){1-5} \cmidrule(l){8-9} 
\multirow{2}{*}{AntifakePrompt} & MS COCO vs. SD2 & 4.86K & \bf97.90 & 96.70 & 93.00 & 91.57 & \multicolumn{2}{c}{91.59} \\
 & MS COCO vs. SD2+LaMa & 4.86K & 97.77 & \bf97.20 & \bf97.10 & \bf93.63 & \multicolumn{2}{c}{\bf92.60} \\
\multirow{2}{*}{Bi-LORA} & MS COCO vs. SD2 & 5.24M & 81.67 & 34.37 & 47.33 & 33.00 & \multicolumn{2}{c}{81.05} \\
 & MS COCO vs. SD2+LaMa & 5.24M & 87.37 & 88.50 & 89.73 & 74.23 & \multicolumn{2}{c}{91.09} \\ \bottomrule
\end{tabular}

\end{adjustbox}
\end{table*}

To this end, we are conducting a comprehensive evaluation of our approach on another distinct dataset, presenting different challenges and characteristics. The data set selected for this evaluation has been chosen to represent varying degrees of divergence from the LSUN bedroom dataset, ensuring a rigorous examination of the versatility and adaptability of our approach.

Table \ref{tab:TrainedOnCvprDataResults} presents the performance of two leading methods, AntifakePrompt and Bi-LORA, on a variety of data sets. These datasets encompass a variety of domains, including image captions (MS COCO), social media (Flickr), synthetic images from text-image generators (SD2, SDXL, IF, DALLE-2, SGXL), image stylization (ControlNet), image inpainting (LaMa, SD2-Inpainting), super-resolution (LTE, SD2-SuperResolution), as well as deep forensic analysis and attack scenarios (Deeper Forensics, Adversarial (Adver.), backdoor attacks, and data poisoning).

AntifakePrompt and Bi-LORA were trained on different sets of data and evaluated on these diverse data sets. The results highlight their effectiveness in the synthetic image detection task. Notably, both methods achieve competitive performance on most data sets, demonstrating their adaptability in real-world applications. Analyzing the table, it's evident that AntifakePrompt consistently achieves high accuracy across several data sets, particularly when trained on MS COCO vs SD2+LaMa. This indicates its robustness in handling diverse data distributions and adversarial scenarios. On the other hand, Bi-LORA exhibits slightly lower performance in some cases, but still maintains strong results across various data sets.

Moreover, the comparison between training on different datasets (e.g., MS COCO vs. SD2) provides valuable insights into the impact of training data composition on model performance. For example, training on a combination of MS COCO and SD2+LaMa data leads to improved performance in most cases, highlighting the importance of incorporating diverse training data to improve model robustness.

\section{Conclusion}
\label{sec:conclusion}

In this paper, we introduced Bi-LORA, a novel approach to synthetic image detection in response to advancements in realistic image generation. We reconceptualized binary classification as an image captioning task, leveraging the powerful convergence between vision and language, as well as the zero-shot nature of \acs{vlm}s. Obtained results demonstrated a remarkable average accuracy of \textbf{93.41\%} in the detection of synthetic image, underling the relevance and effectiveness of Bi-LORA approach to the challenges posed by images generated by unknown generative models. Moreover, unlike previous studies that require millions of parameters to be tuned/learned, Bi-LORA model only needs to adjusts far fewer parameters, thus setting a better balance between training cost and efficiency.

Future research could be conducted on the possibility of distilling the mixture of knowledge from very large \acp{vlm} to a lightweight model, given that our experimental results highlight the complementarity between Bi-LORA and VitGPT2, considering that \acp{vlm} require significant computational and memory resources. Moreover, we aim to take a step further beyond just detecting synthetic images from real ones by attributing them to their source generator using \acp{vlm} in a multitask single-step approach, thereby holding generators accountable for their actions in today’s rapidly evolving technological landscape.

To support the principle of reproducible research and to enable future extensions, we make the code and models publicly available at \href{https://github.com/Mamadou-Keita/VLM-DETECT}{https://github.com/Mamadou-Keita/VLM-DETECT}.

% In this paper, we have presented a new approach to the synthetic image detection task by exploiting the capabilities of state-of-the-art \acp{vlm}. By re-framing the challenge of binary classification as an image captioning task, we showed the potential of our proposed Bi-LORA model to transcend traditional boundaries and provide new perspectives in image authenticity assessment. Our methodology relies on the ability of these models to fuse visual and textual information, enabling them to generate captions that encapsulate the essence of both real and synthetic images. The results of our experiments have shown that this approach outperforms conventional classification techniques.  Our results have implications beyond the immediate task of image classification \hl{Here give some number of detection and generalisation for distortion and GAN}. They highlight the potential of \acp{vlm} to solve complex problems that benefit from multimodal understanding. As these models evolve, the boundaries of their applications expand, promising to revolutionize various fields that rely on nuanced perception.

% Our evaluation provides in-depth insights into the capabilities of detectors, aimed at assessing the performance and effectiveness of these architectures in the task of detecting images generated by artificial intelligence

\section*{Acknowledgments}
This work has been funded by the project PCI2022-134990-2 (MARTINI) of the CHISTERA IV Cofund 2021 program.
%, funded by MCIN/AEI/10.13039/501100011033 and by the “European Union NextGenerationEU/PRTR”; by the research project DisTrack: Tracking disinformation in Online Social Networks through Deep Natural Language Processing, granted by Mobile World Capital Foundation; by the Spanish Ministry of Science and Innovation under FightDIS (PID2020-117263GB-I00); by MCIN/AEI/10.13039/501100011033/ and European Union NextGenerationEU/PRTR for XAI-Disinfodemics (PLEC2021-007681) grant, by Comunidad Autónoma de Madrid under S2018/TCS-4566 grant, by European Comission under IBERIFIER - Iberian Digital Media Research and Fact-Checking Hub (2020-EU-IA-0252); and by "Convenio Plurianual with the Universidad Politécnica de Madrid in the actuation line of Programa de Excelencia para el Profesorado Universitario".

%\clearpage

\bibliographystyle{IEEEtran}
\bibliography{references}

\end{document}